%% file: main.tex
\documentclass[conference]{IEEEtran}
\usepackage{times}

\usepackage[numbers]{natbib}
\usepackage{multicol}

\usepackage[svgnames]{xcolor}

\definecolor{myblue}{rgb}{0.21,0.49,0.74}
\usepackage[pagebackref,breaklinks,colorlinks,citecolor=myblue]{hyperref}

\IEEEoverridecommandlockouts  
\input{commands.tex}

\usepackage{enumitem}
\usepackage{xspace}
\newcommand{\alias}{RoboScript\xspace}
\def \robotlogo {\raisebox{-0.1\height}{\includegraphics[height=0.95\baselineskip]{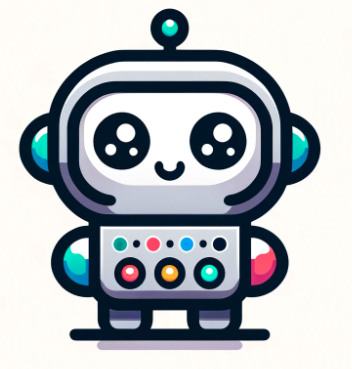}}}

\begin{document}

% paper title
\title{\robotlogo{}\textbf{R}obo\textbf{S}cript: Code Generation for Free-Form Manipulation Tasks across \textbf{R}eal and \textbf{S}imulation}

\author{
\small
    Junting Chen$^{*12}$, Yao Mu$^{*13}$, Qiaojun Yu$^{4}$, Tianming Wei$^{4}$, Silang Wu$^{5}$, Zhecheng Yuan$^{5}$, Zhixuan Liang$^{3}$, 
    \\
    Chao Yang$^{1}$, Kaipeng Zhang$^{1}$, Wenqi Shao$^{1}$, Yu Qiao$^{1}$, Huazhe Xu$^{5}$, Mingyu Ding$^{\dagger6}$, Ping Luo$^{\dagger13}$
    \vspace{0.2cm} \\ 
    OpenGVLab, Shanghai AI Laboratory$^{1}$ \quad ETH Zurich$^{2}$ \quad The University of Hong Kong$^{3}$ 
    \\
    Shanghai Jiao Tong University$^{4}$ \quad Tsinghua University$^{5}$ \quad UC Berkeley$^{6}$
\\
    \thanks{
    \small
This work was completed during the internship at the OpenGV Lab, Shanghai AI Laboratory by Yao Mu and Junting Chen. Mingyu Ding and Ping Luo are the co-corresponding authors.}\\
        \thanks{* Equal contribution.\ \ \ \ $\dagger$ Corresponding Authors.}\\
}

\maketitle

\input{sec/0_abstract}

\IEEEpeerreviewmaketitle

\input{sec/1_introductions}

\input{sec/2_relworks}

\input{sec/3_pipeline.tex}

\input{sec/4_benchmark.tex}

\input{sec/5_experiments}

\input{sec/6_conclusions}

\section*{Acknowledgments}
We would like to express our sincere appreciation to Professor Yufeng Yue Mr. Guangyan Chen from the Beijing Institute of Technology for their great help in code deployment onto real robots, and their valuable feedback for the user-friendly development of the project. Equally, our gratitude extends to Professor Cewu Lu and Dr. Wenhai Liu from Shanghai Jiao Tong University, for their great support in helping integrate the Anygrasp model into our system.

%% Use plainnat to work nicely with natbib. 

\bibliographystyle{plainnat}
\bibliography{main}

\clearpage
\input{sec/7_appendix}
\end{document}

%% file: commands.tex
\usepackage{verbatim}

\newcommand{\eg}{\emph{e.g.}\xspace}
\newcommand{\ie}{\emph{i.e.}\xspace}

\definecolor{orange}{rgb}{1,0.5,0}
\definecolor{maroon}{rgb}{0.51,0,0}

\usepackage{times}
\usepackage{epsfig}
\usepackage{graphicx}
\usepackage{amsmath}
\usepackage{amssymb}
\usepackage{xspace}
\usepackage[font=small]{caption}
\usepackage[linesnumbered,ruled,vlined]{algorithm2e}

\usepackage{subcaption}
\usepackage[font=small]{caption}
\usepackage{multirow}

\usepackage[labelformat=simple]{subcaption}

\usepackage[flushleft]{threeparttable}

\usepackage[utf8]{inputenc} % allow utf-8 input
\usepackage[T1]{fontenc}    % use 8-bit T1 fonts
\usepackage{url}            % simple URL typesetting
\usepackage{booktabs}       % professional-quality tables
\usepackage{amsfonts}       % blackboard math symbols
\usepackage{nicefrac}       % compact symbols for 1/2, etc.
\usepackage{microtype}      % microtypography
\usepackage{mathtools}
\usepackage{bbm}
\usepackage{bm}
\usepackage{braket}
\usepackage{xspace}

\usepackage{subcaption} 
\usepackage{caption}
\usepackage{wrapfig}

\usepackage{amsthm,commath}
\usepackage[utf8]{inputenc}
\usepackage[english]{babel}

\newcommand{\supp}{{Appendix}\xspace}

\theoremstyle{plain}

\theoremstyle{definition}

\theoremstyle{remark}

\usepackage{xspace}
\usepackage{comment}
\usepackage{balance}
\usepackage{etoolbox,siunitx}
\usepackage{calc}
\usepackage{pifont,hologo}

\usepackage[super]{nth}
\usepackage{nicefrac}
\def\tt{\mathbf{t}}

\DeclareMathSymbol{@}{\mathord}{letters}{"3B}

\def\latex/{\LaTeX}
\def\bibtex/{\hologo{BibTeX}}

\usepackage[official]{eurosym}

\SetKwComment{Comment}{/* }{ */}

\usepackage[hmargin=2cm,vmargin=2.5cm]{geometry}
\usepackage{etoolbox}
\usepackage{hhline}
\usepackage{tablefootnote}
\usepackage{pdflscape}
\usepackage{colortbl}%

\usepackage{booktabs}

\newif\ifblackandwhite
\ifblackandwhite
  
\else
\fi

\usepackage[linesnumbered,ruled,vlined]{algorithm2e}
\usepackage{listings}
\usepackage{subcaption}
\newcommand{\algname}[1]{\textsc{#1}}

\renewcommand{\vec}[1]{\mathbf{#1}} 
\newcommand{\mat}[1]{\mathbf{#1}} 

\usepackage{tcolorbox}
\usepackage{etoolbox}
\usepackage[frozencache=true,cachedir=.]{minted} 

\definecolor{bg}{rgb}{0.95,0.95,0.95}
\usepackage[numbers]{natbib}
\usepackage{multicol}
\usepackage{adjustbox}

\newcommand{\bx}{\boldsymbol{x}}
\newcommand{\by}{\boldsymbol{y}}
\newcommand{\bz}{\boldsymbol{z}}

\newcounter{codeBlockCounter}
\makeatletter
\newcommand{\labelText}[2]{%
#1\refstepcounter{codeBlockCounter}%
\immediate\write\@auxout{%
  \string\newlabel{#2}{{1}{\thepage}{{\unexpanded{#1}}}{codeBlockCounter.\number\value{codeBlockCounter}}{}}%
}%
}
\makeatother

\usepackage{todonotes}

%% file: sec/0_abstract.tex
\begin{abstract}
Rapid progress in high-level task planning and code generation for open-world robot manipulation has been witnessed in Embodied AI.
However, previous studies put much effort into general common sense reasoning and task planning capabilities of large-scale language or multi-modal models, relatively little effort on ensuring the deployability of generated code on real robots, and other fundamental components of autonomous robot systems including robot perception, motion planning, and control.
To bridge this ``ideal-to-real'' gap, this paper presents \textbf{RobotScript}, a platform for 1) a deployable robot manipulation pipeline powered by code generation; and 2) a code generation benchmark for robot manipulation tasks in free-form natural language.
The RobotScript platform addresses this gap by emphasizing the unified interface with both simulation and real robots, based on abstraction from the Robot Operating System (ROS), ensuring syntax compliance and simulation validation with Gazebo. We demonstrate the adaptability of our code generation framework across multiple robot embodiments, including the Franka and UR5 robot arms, and multiple grippers. Additionally, our benchmark assesses reasoning abilities for physical space and constraints, highlighting the differences between GPT-3.5, GPT-4, and Gemini in handling complex physical interactions. Finally, we present a thorough evaluation on the whole system, exploring how each module in the pipeline: code generation, perception, motion planning, and even object geometric properties, impact the overall performance of the system. 
\end{abstract}

%% file: sec/1_introductions.tex
\section{Introduction}
\label{sec:intro}

% TODO: Previous works, text-based high-level only.

\begin{figure*}
    \centering
    \vspace{-10pt}
    \includegraphics[width=0.95\linewidth]{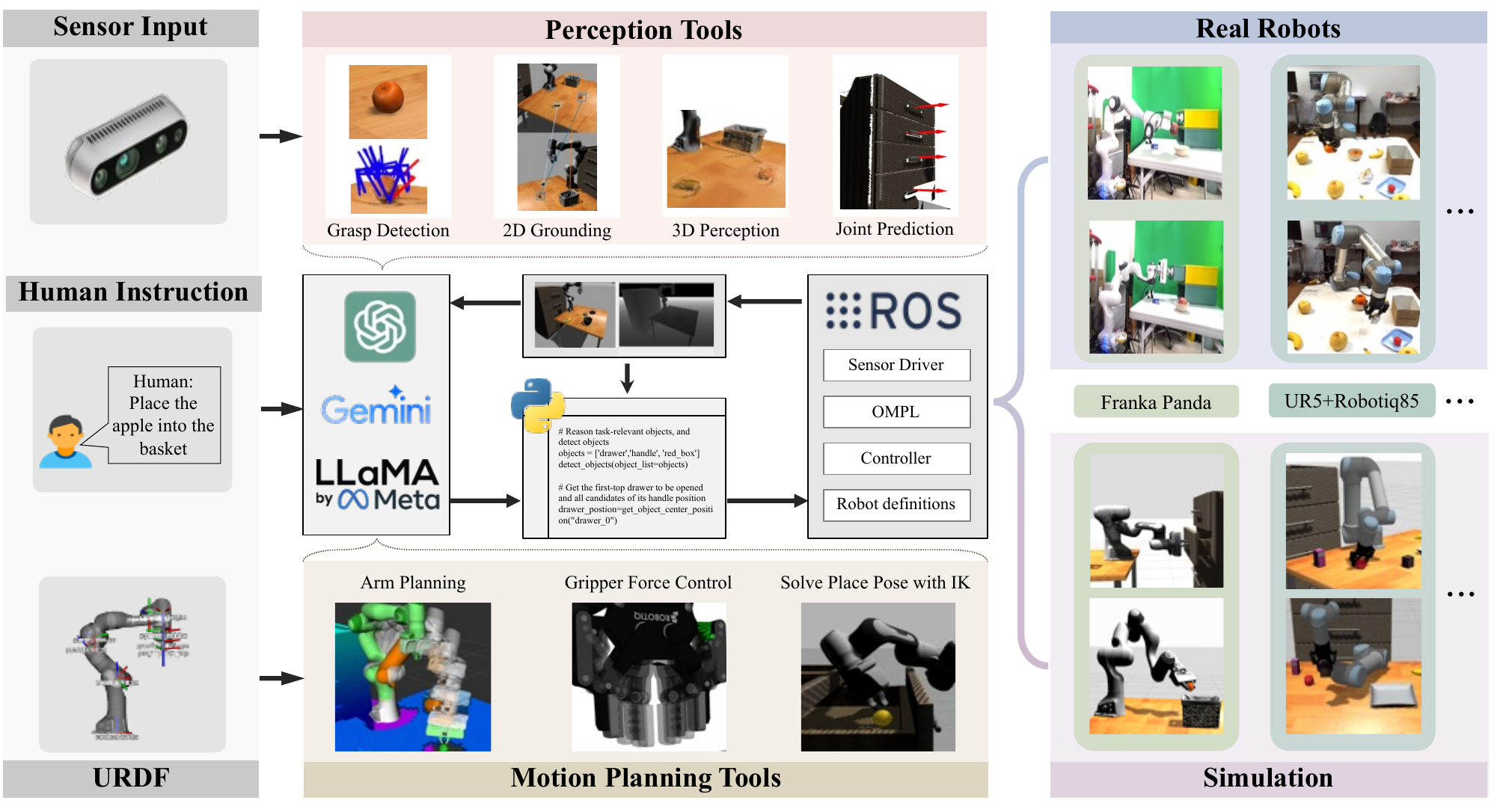}
    \caption{\textbf{Framework of \alias} . The input layer contains sensor input, human instruction, and robotic URDF (Unified Robot Description Format) data. The system utilizes various perception tools, such as grasp detection, 2D grounding, 3D perception, and joint prediction, to interpret the input data. These tools are integrated with motion planning tools that include arm planning, gripper force control, and solving place pose with inverse kinematics (IK). The Robot Operating System (ROS) serves as the middleware to provide abstraction to sensor drivers, controllers, and robot definitions across real robots and the simulation. The framework controls multiple real robots and their counterparts in the simulation with a unified code generation pipeline. This modular approach enables flexibility in robotic applications and adaptability to new code generation methods or robot architectures, from simple tasks to complex manipulations.}
    % \vspace{-10pt}
    \label{fig:enter-label}
    \vspace{-10pt}
\end{figure*}

Leveraging pre-trained language models for robotic applications is an active research area, with many works focusing on planning and reasoning ~\cite{huang2022language,zeng2022socratic,chen2022open,shah2022lm,huang2022inner,singh2022progprompt,huang2022visual,raman2022planning,song2022llm,vemprala2023chatgpt,lin2023text2motion,ding2023task,huang2023grounded,driess2023palm,yuan2023plan4mc,xie2023translating,lu2023multimodal,patel2023pretrained,jansen2020visually,wang2023voyager,yang2023pave,wang2023gensim}. To enable language models to perceive physical environments, textual scene descriptions ~\cite{huang2022inner,zeng2022socratic,singh2022progprompt}  or perception tools ~\cite{liang2023code}  are provided.  Integrating large language models (LLMs) with robots has significantly advanced robotics, enhancing decision-making and control through improved language understanding and task execution. However, current benchmarks focus more on high-level semantic understanding, often overlooking the nuances of lower-level control and physical constraints in robotic manipulation.

To address this gap, we introduce the RobotScript Benchmark, which maps human language instructions to robot motion planning and provides evaluation within physically realistic contexts. Going beyond traditional semantic parsing, it incorporates nuances of physical interactions, workspace constraints, and object properties that are critical for real-world robotic applications. By assessing both conceptual and embodied intelligence, RobotScript combines the rich semantic and pragmatic understanding of large language models with the fine-grained motor control needed in dynamic, real-world settings. The designed system provides a complete pipeline from 2D image detection to 3D scene modeling, grasp pose prediction, and finally, motion planning. This enables the robot to both understand high-level natural language commands and autonomously leverage various perception tools and planning algorithms. As a result, it can generate low-level motion control, without the need for supervised information.

The RobotScript Benchmark revolves around three core components:
\textbf{1) ROS-based Code Generation Testing}: The proposed benchmark facilitates testing and validation of generated code within the Robot Operating System (ROS) framework, enabling direct deployment on physical robots. It supports connectivity with various sensors, actuators, and large neural network models, ensuring the syntactic validity of generated code and simulation-based testing in Gazebo. To demonstrate the versatility across robotic platforms, we evaluated our framework on different robot embodiments, assessing the impact of mechanical designs and gripper types on task performance. Historically, the incompatibility between the ROS and conda environments has impeded the deployment and testing of AI models on robotic hardware. By leveraging recent advancements in  Robostack~\cite{FischerRAM2021}, which is a conda-compatible ROS infrastructure, our benchmark enables seamless model implementation within any conda environment.
%
% ROS-Based Code Generation Testing: The benchmark facilitates the testing of generated code within the ROS framework, allowing for direct deployment on actual robots. It supports various sensors, actuators, and large model network communications, ensuring code syntax compliance and Gazebo-based simulation validation. We further validate our framework's versatility across different robot embodiments, assessing the impact of mechanical structures and gripper forms on task success. \junting{In the past, the mutually exclusive nature of the ROS and conda environments greatly hindered the deployment and testing of AI models on real robotic systems. Our benchmark, powered by latest conda-compatible ROS infrastructure, Robostack [citation here], allows for model implementation within arbitrary conda environment.}
%
\textbf{2) Perception-in-the-loop Benchmark Pipeline}: In order to make our tests more closely resemble real robot scenarios, in our benchmark, the inputs of both the robot planner and controller are based on the results from robot perception. We provide multiple tools, such as AnyGrasp~\cite{fang2023anygrasp}, to predict a grasp pose from a segmented object point cloud which is constructed by multi-view 3D fusion, and then used as the input to the grasp motion planning API. Basing the motion planning on the perceived grasp pose, rather than ground truth, introduces realism and potential errors that would occur in a live system. However, we also supply the original ground truth RGBD images and 3D bounding box data to mitigate excessive noise from potential perception failures. This pipeline is directly deployable to real robots by using sensor RGBD images and predicted bounding boxes.
\textbf{3) Physical Space and Constraint Reasoning}: Our benchmark introduces a comprehensive testbed that evaluates reasoning capabilities regarding physical space and constraints. These tests involve understanding complex interactions among various objects, such as calculating feasible transitional positions without causing interference with other objects. The benchmark underscores the challenges in constraint satisfaction by highlighting the performance differences between GPT-3.5 and GPT-4 in these scenarios.

To summarize, the contributions of this paper include: 
% \vspace{-10pt}
\begin{itemize}[leftmargin=9pt] % itemsep=0pt
  \item \textbf{Comprehensive Integration of LLMs with Robotics}: RoboScript effectively bridges the gap between high-level semantic understanding, as well as the practical nuances of lower-level control and physical constraints in robotic manipulation. It provides an autonomous manipulation pipeline covering task interpretation, object detection, pose estimation, grasp planning, and motion planning.
  % The benchmark implements an autonomous pipeline for robotic manipulation spanning from semantic understanding of tasks, to object detection, pose estimation, grasp planning, and finally motion planning for execution.
  
\item \textbf{In-depth Ablation Study on System Modules}: We provide an ablation study on each system module, deeply analyzing the impact of individual module errors. This enhances the benchmark's relevance and practicality in real-world robotic applications.

\item \textbf{Assessment of LLM Reasoning for Physical Interactions}: Our benchmark assesses reasoning abilities for physical space and constraints, highlighting the differences between GPT-3.5, GPT-4, and Gemini in handling complex physical interactions.
    % \item This paper presents a general real-time robotic system for manipulation purposes by leveraging the high-level task planning and reasoning capability of Large Language Models (LLMs), the abstraction of 

    % This paper presents the RobotScript Benchmark, a comprehensive platform to evaluate code generation in the context of physical reality. The benchmark incorporates the nuances of physical interactions, workspace constraints, and object properties, all of which are crucial for real-world robotic applications. The benchmark consists of three core components: ROS-Based Code Generation Testing, Perception-in-the-loop benchmark pipeline, and Physical Space and Constraint Reasoning. By integrating high-level cognitive abilities offered by Large Language Models (LLMs) with the intricate requirements of real-world physical control in robotic manipulation, the RobotScript Benchmark provides an evaluation platform that goes beyond traditional semantic parsing. Overall, this paper offers a significant contribution to the field of robotics, facilitating the development of foundation models that can work across different robotic platforms.
\end{itemize}

%% file: sec/2_relworks.tex
\section{Related Work}
\label{sec:rel}

% \section{Related Work}

\subsection{Large Language Models for Robotics}
Recent research has delved into the integration of large language models (LLMs) with embodied AI tasks, focusing on planning, reasoning, and control~\cite{radford2019language, brown2020language}. A common approach involves feeding LLMs with perceptual inputs, such as scene descriptions~\cite{huang2022inner, zeng2022socratic} or visual information~\cite{driess2023palm, openai2023gpt}, and implementing action capabilities through libraries of motion primitives~\cite{liang2023code}. Despite these promising developments, effectively representing intricate spatial relationships continues to be a formidable challenge for LLMs.
Several works have demonstrated the potential of LLMs in composing sequential policies for embodied domains including VoxPoser~\cite{voxposer} for robotic manipulation, SayCan~\cite{saycan} for instruction following, Palm-E~\cite{palme} for dexterous manipulation, RT-2~\cite{rt2} for mobile manipulation, EmbodiedGPT~\cite{embodiedgpt} for motion planning, Voyager~\cite{wang2023voyager} and Smallville~\cite{park2023generative} for embodied gaming, VisProg~\cite{gupta2023visual} for program synthesis, TAPA~\cite{tapa} and SayPlan~\cite{rana2023sayplan} for task planning.
While significant strides have been taken, fully closing the perception-action loop in embodied agents remains a dynamic and ongoing area of research. Key challenges encompass representing complex spatial relationships and long-term planning, enhancing sample efficiency in learning, and seamlessly integrating perception, reasoning, and control within a unified framework. Further exploration of benchmark tasks, and physical knowledge representation will propel LLMs towards achieving generalized embodied intelligence.

\subsection{Benchmarks for Robotic Code Generation}
Ravens~\cite{zeng2021transporter} was originally proposed as a simulation benchmark for reinforcement learning of robotic manipulation skills.
% consisting of 6 tabletop rearrangement tasks using a simulated UR5 arm with a suction gripper in PyBullet. Object placements are randomized to evaluate agents' generalization ability. 
It was later extended by CLIPort~\cite{shridhar2022cliport} to include language conditioning, allowing for the study of language-conditioned policy learning and generalization to new instructions.
% The CLIPort~\cite{shridhar2022cliport} extended Ravens to 10 tasks by adding language conditioning, where natural language instructions specify the goal description instead of goal images. This enables to study language-conditioned policy learning and generalization to new instructions. CLIPort uses CLIP~\cite{clip} for converting instructions to goal images.
Building on the Ravens benchmark, RoboCodeGen~\cite{liang2023code} was developed to test language models to test language models on robotics programming problems.
% It consists of 37 function generation tasks like finding the closest point and checking bounding box containment. 
RoboGen~\cite{wang2023robogen} further expanded this by encouraging the use of libraries like NumPy and allowing for the definition of helper functions. These benchmarks have provided valuable platforms for evaluating the performance of various Large Language Models (LLMs) (\eg, RoboCodeGen supports evaluation on GPT-3, InstructGPT and Codex) in generating code for robotic tasks.

% RoboGen~\cite{wang2023robogen} encourages using libraries like NumPy and allows defining helper functions. 
% RoboCodeGen supports evaluation on 4 LLMs - two general language models (GPT-3, InstructGPT) and two code-focused models (Codex). RoboGen provides a way to benchmark code generation models on non-trivial robotics reasoning tasks.

\subsection{Motion Planning}
In the realm of robot motion planning, the Open Motion Planning Library (OMPL)~\cite{sucan2012open} is a key open-source framework that offers a range of advanced algorithms for navigating complex environments. MoveIt~\cite{Coleman2014ReducingTB} was developed to integrate OMPL's capabilities with the ROS ecosystem, providing a comprehensive platform that includes SOTA motion planning, manipulation, 3D perception, kinematics, and control. This integration greatly improves the applicability and efficiency of robot motion planning within the ROS ecosystem. 
% In the field of motion planning for robotic arms, OMPL is developed as an open-source motion planning library, and contains implementations of many complex planning algorithms. To simplify and facilitate the use of OMPL in ROS, MoveIt!\cite{Coleman2014ReducingTB} is created. MoveIt! is an integrated development platform incorporating the latest advances including motion planning, manipulation, 3D perception, kinematics, and control. 
\subsection{Grasp Pose Detection}
Vision-guided grasp pose detection is a crucial area in robotics research, transitioning from 4 degrees of freedom (DOF) top-down grasping methods~\cite{paper21, paper23, paper27} to more complex 6 DOF grasping methods~\cite{breyer2021volumetric, jiang2021synergies}. The state-of-the-art 6 DOF grasp pose detection model, AnyGrasp~\cite{fang2023anygrasp}, achieves a success rate in object grasping comparable to human performance by studying the geometric features of objects. This has paved the way for further work on specified object grasping~\cite{liu2024ok, ju2024robo} or articulated object manipulation~\cite{yu2023gamma}.

%% file: sec/3_pipeline.tex
\section{RoboScript Pipeline}
\label{sec:roboscript-pipeline}

In this section, we introduce the RoboScript pipeline, which is designed to deploy and evaluate Large Language Models (LLMs) in generating code to control robotic arms for completing embodied AI tasks as instructed by humans. Given RGBD images from the sensor and human instructions as input, the pipeline generates Python code that directly manipulates a robotic arm. By leveraging the capabilities of infrastructure including 1) the highly versatile communication protocols and message interface provided by the Robot Operating System (ROS), 2) the ROS-integrated Gazebo simulation suite, and 3) the versatile motion planning framework MoveIt, the pipeline creates a unified interface to both the simulation and real robots. This abstraction allows our pipeline to have a snippet of generated code tested safely in a simulated environment and then executed in reality.

\subsection{Pipeline Overview}
\begin{figure*}
    \centering
    \includegraphics[width=0.95\linewidth, trim={0cm 0cm 0cm 0cm}, clip]{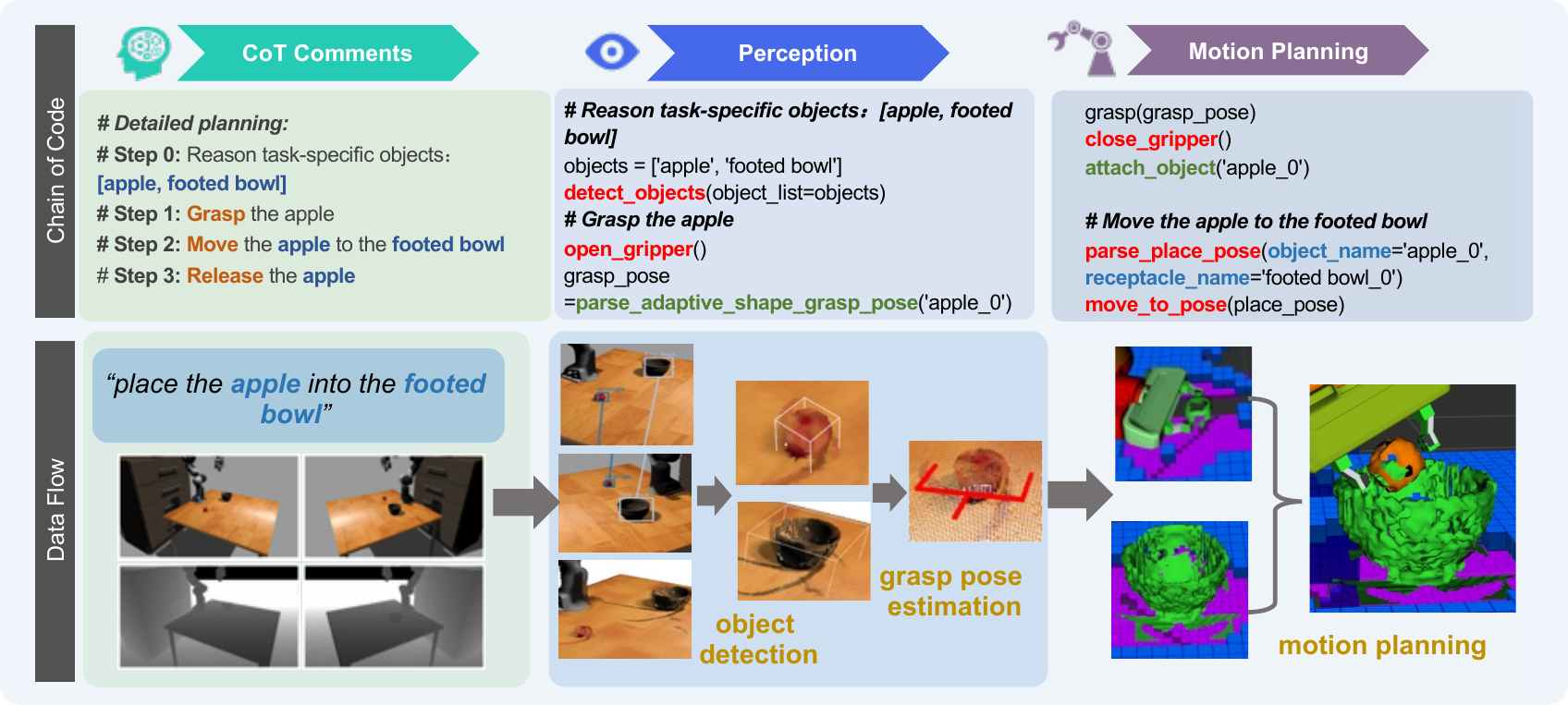 }
    \vspace{2pt}
    \caption{\textbf{Pipeline of \alias}. \alias uses Perception and Motion Planning Tools, activated by a task query, to generate a Python script with the LLM. This script includes comments, processes images into 3D models, and plans safe robot movements. Perception Tools identify objects and spatial details for planning. Motion Planning Tools then create a collision-free path for actions.}
    \label{fig:pipeline}
    \vspace{-10pt}
\end{figure*}

As described in the introduction, our framework provides a set of Perception Tools and Motion Planning Tools. Each time a task query is sent to the system, the LLM generates a Python script by using In-context Learning (ICL)~\cite{dong2022survey}. This means that the few-shot example codes are integrated into the prompt as demonstrations. Then, the code is executed to finish the task, as shown in Fig.~\ref{fig:pipeline}. 

A generated Python script mainly consists of three parts: 
\begin{enumerate}
    \item \textbf{Chain-of-Thought (CoT)~\cite{wei2022chain} Comments} help LLMs decompose long-time-horizon and complex tasks.
    \item \textbf{Perception Tools} process the raw sensor input of multi-view RGB-Depth images to 3D representations and spatial arguments. When the ``\textit{detect\_obejects}'' function is called, the system perception pipeline is triggered, building 3D representations and implicitly sending them to the Moveit planning scene. This behavior is designed to support fundamental functionalities of robotic system like obstacle-aware path planning. Other perception utils are then called to get spatial arguments like positions and orientations for motion planning.
    \item \textbf{Motion Planning Tools} take the output of perception utils as input arguments, generate a collision-free path plan, and further call low-level controllers to control the motion. For example, when opening a drawer attached to a cabinet, the robot needs to pull the drawer handle with a Cartesian path outward. The path should be parallel to the prismatic joint between the drawer and the cabinet. The motion plan generates the path based on the predicted prismatic joint direction from the perception tool ``\textit{get\_object\_joint\_info}''.
\end{enumerate}
 You can find the details of all our tools API, their explanation, their arguments, and returns definitions in \supp \ref{sec:appendix_tools}.

It is also worth noting that there could be multiple rounds of perception $\rightarrow$ motion planning in the code script, especially when the task requires multiple rounds of robot-environment interaction. For instance, consider a fundamental task for (future) household robots: storing table containers in a cabinet. The robot might first detect the environment and find a closed cabinet, with no direct motion plan to place a container into the drawer. In this case, the robot needs to first interact with the environment by opening the drawer and retracting its arm. It then needs to re-detect the scene to update its plan.

\subsection{Code Generation}
\label{pipeline: code_generation}
The full prompt for code generation involves three sections: 
\begin{enumerate}
    \item \textbf{System Prompt}: A system prompt is a pre-defined and query-irrelevant input that sets the context and general instructions for the LLM to respond. It helps structure the output of long texts. GPTs and Llama-series models are trained with special data to increase the model's attention on system messages wrapped in special tokens \cite{OpenAI2023GPT4TR,touvron2023llama}. We provide all format-related instructions, e.g., ``\textit{You should only generate Python code and comments.}'' in this section. Since LLMs generally pay better attention to the system prompt, especially when reading long text, we also include descriptions/doc-strings of perception and motion planning tools into the system prompt.
    \item \textbf{Few-shot Examples}: To limit the LLM output format and help reduce hallucination, we provide 10 rounds of {task query: master script code} dialogues so that LLMs can learn from demonstrations how to respond to a task query input. % \textcolor{red}{Should we also show some examples here?}
    \item \textbf{Task Query}: The task query consists of two sentences. The first sentence describes what objects are present in the current environment. The second sentence provides the task instruction. For example, "objects=[apple, bowl]. Please place the apple and place it into the bowl" is a valid task query. For LLMs, the user needs to input the scene description. For Large Multimodal Models (LMMs), the scene description could also be generated by querying the LMM with an RGB image and another pre-defined query prompt.
\end{enumerate}

Similar to \citet{liang2023code}, \alias also employs a hierarchical agent architecture to facilitate modular code generation. At the top is the main agent responsible for analyzing the problem and producing the overall workflow. As the main agent generates code, a syntax parser operates in parallel, continuously verifying the syntactic correctness of each function call. When the parser identifies a function that is not in loaded variables, it triggers a specialized sub-agent to generate the required function from the function name and description. The sub-agent then produces Python function definitions as requested, using a similar three-section prompt as model input. 

However, even for the state-of-the-art GPT-4 model, this hierarchical decomposition is not sufficient when generating code for tasks with long time-horizons or tasks involving complex reasoning processes. Chain-of-Thought \cite{wei2022chain} (CoT), a prompting method, is adopted to teach the LLM to explicitly print its thinking process step by step, as shown in the ``CoT Comments'' part of Fig. ~\ref{fig:pipeline}. By leveraging the power of contextual attention in auto-regressive models, CoT significantly enhances the reasoning capability of LLMs. Many following works and variants \cite{yao2023tree,yao2023beyond} 
% have demonstrated that this explicit thinking paradigm is powerful for long-time-horizon and complex tasks. 
have once again validated its advantages.
Viewing our pipeline as a baseline for future embodied control code generation architectures, we accept the original CoT prompting and find that CoT already performs well on the benchmark tasks. 

% \TODO{Add the experiment result with and without CoT comments}
\vspace{-2pt}
\subsection{Perception Pipeline}

In this section, we introduce the perception pipeline, a key component of the Roboscript system. As mentioned in the overview, ``\textit{detect\_objects}'' function, which serves as the foundation and prerequisite of all the other perception tools, triggers the perception pipeline of the system, conducts 2D visual grounding, builds 3D representations, and updates the Moveit planning space with reconstructed meshes. The LLM can then choose the perception tools API to retrieve the result of 3D representations or further process it for more advanced information. For example, ``\textit{get\_3d\_bbox}'' simply returns the stored object instance 3D bounding box, while ``\textit{get\_plane\_normal}'' returns the surface normal vector of the plane closest to the given position. This design is motivated by the observation that 1) 3D reconstruction and object 3D bounding boxes, with relatively fixed algorithms to calculate, are shared as the foundation data of most other perception tools, and 2) Moveit's planning scene, which is the data infrastructure of motion planning tools, is organized based on objects' meshes and an Octomap~\cite{hornung2013octomap} for collision-free planning. Therefore, we encapsulate these low-level perception processes in a high-level API, This approach effectively minimizes code length and reduces redundancy, also relieving the difficulties encountered when generating lengthy texts with language models. 

For clarity, we define all the mathematical notations in the perception pipeline as follows: Generally, we assume the robot system is equipped with $k\geq1$ RGB-Depth cameras, generating images $\mat{I}^{0}, \dots , \mat{I}^{k-1}$.  $\mat{I}^{i}$ denotes the input RGBD image from camera $i$, $\mat{I}^i_{rgb}\in\mathbb{R}^{H\times W\times 3}$ denotes the input RGB image of height $H$ and width $W$, and $\mat{I}^{i}_{depth}\in\mathbb{R}^{H\times W}$ denotes the depth image. For camera $i$, denote the camera intrinsic matrix as $\mat{K_i}\in\mathbb{R}^{3\times3}$. The rotation and translation of the camera in the world frame are represented as $\mat{R_i}\in\mathbb{R}^{3\times3}$ and $\vec{t_i}\in\mathbb{R}^{3}$, respectively. A 2D bounding box on $i$-th RGB image $\mat{I}^{i}_{rgb}$ with label $o_j$ is denoted as $\vec{b}^{ij}_{2D}=\left[\bx_{min}, \by_{min}, \bx_{max}, \by_{max}\right]$. And a 3D bounding box of object $o_j$ is denoted as $\vec{b}^{j}_{3D}=\left[\bx_{min}, \by_{min}, \bz_{min}, \bx_{max}, \by_{max}, \bz_{max}\right]$. A 3D point cloud is denoted as $\mat{P}=\{\vec{p}_0, \dots ,\vec{p}_n\}$, where $\vec{p}_i=[\bx_i, \by_i, \bz_i]\in\mathbb{R}^3$ represents a point in space. A grasp pose is a 3D-homogeneous (\ie SE(3)) transformation denoted as $\mat{T}\in\mathbb{R}^{4\times4}$. Since we build the whole system on top of \citet{sucan2012open}, we have chosen to omit the details of joint motion planning and focus solely on the end-effector pose in code generation. Thus, in the rest of this paper, we use the term ``motion trajectory'' to indicate a sequence of waypoints of end-effector poses. We denote a motion trajectory as $\mat{M}=\{\mat{T}_0, \dots ,\mat{T}_n\}$, where $\mat{T}_i\in\mathbb{R}^{4\times4}$ represents an end-effector pose.

\subsubsection{\textbf{2D Grounding}}
By leveraging the capability of the open-set text-to-image grounding model GLIP \cite{li2021grounded}, the pipeline first detects a list of task-relevant objects, reasoned by the LLMs with scene description and task instruction, on each RGB image $\mat{I}^{i}_{rgb}$. 
GLIP generates a list of 2D bounding boxes $\vec{b}^{ij}_{2D}$ on each RGB image $\mat{I}^{i}_{rgb}$ of object label $o_j$, given its text description:
\vspace{-2pt}
\begin{equation}
    \vec{b}^{ij}_{2D} = \text{GLIP}(\mathbf{I}^{i}_{rgb}, \text{Description}(o_j)), i\in\{0, \dots, k-1\}.
\end{equation}

The open-set reasoning capabilities of LLMs, combined with the open-set grounding capabilities of large 2D vision models, empower our system with open-set intelligence capability. Limited by data and training paradigms, 3D models are relatively more constrained to a specific domain or even a dataset. This is the main reason we choose to use a 2D grounding model and manually propagate semantic information to 3D representations, instead of using a 3D grounding model.

\subsubsection{\textbf{3D Reconstruction}}
In an indoor environment with a lot of objects and clutter, varying viewpoints are helpful to circumvent the obstruction issues that can occur with a single view. The multi-view RGBD images $\mat{I}^{0}, \dots , \mat{I}^{k-1}$ are firstly integrated into a Truncated Signed Distance Function (TSDF) volume, using a TSDF fusion pipeline proposed by \citet{zhou2013dense} and implemented in Open3D \cite{zhou2018open3d}. TSDF fusion is often preferred over cloud fusion from depth maps in real-time reconstruction pipelines due to its simplicity, faster speed, and ease of parallelization, as discussed in \cite{sun2021neuralrecon}. This process can be formulated as:
\begin{equation}
\vspace{-4pt}
% \begin{aligned}
    TSDF(\mathbf{x})=\bigoplus_{i=0}^{k-1} F(\mathbf{I}^{i}, \mathbf{x}),
% \end{aligned}
\end{equation}
where $F(\mathbf{I}^{i}, \mathbf{x})$ denotes the fusion process of image $\mathbf{I}^{i}$ at position $\mathbf{x}$, and $\bigoplus$ is the integration operation over all views.

The volumetric representation is further converted to different representations at request: (a) point cloud for grasp detection model AnyGrasp \cite{fang2023anygrasp}, joint prediction model GAMMA, and plane detection model \cite{araujo2020robust}; (b) mesh for Moveit planning scene; (c) uniform voxel grids for GIGA \cite{jiang2021synergies}. The marching Cube \cite{lorensen1998marching} algorithm is used in the surface extraction process to compute point cloud and mesh. 

\subsubsection{\textbf{Cross-view Bounding Box Matching}}
With the TSDF volume and 2D grounding results, which are independent 2D object bounding boxes across images of different camera views, the perception pipeline continues to get a list of 3D object instances by (a) matching 2D bounding boxes across different views and (b) filtering the point cloud within matched 2D bounding boxes. 

\begin{figure}
    \centering
    \includegraphics[width=0.95\linewidth, trim={8cm 1cm 8cm 1cm}, clip]{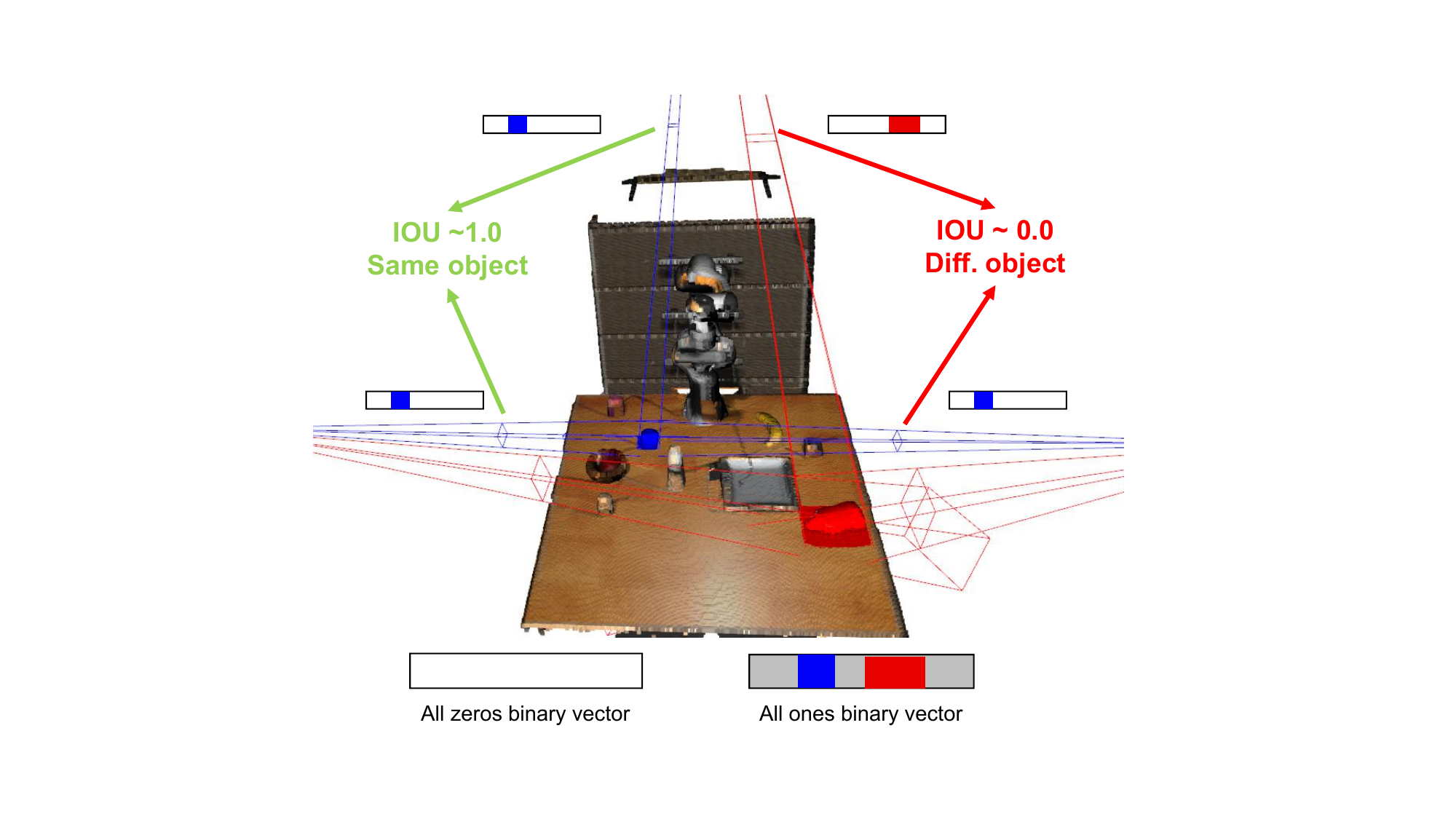}
    \vspace{-10pt}
    \caption{\textbf{Cross-view bounding box matching score computation}. The matching score between two bounding boxes is calculated by the Intersection over Union (IOU) of points presence binary vector.}
    \label{fig:cross_view}
        \vspace{-15pt}
\end{figure}

To match the 2D bounding boxes $\vec{b}^{ij}_{2D}$ on each RGB image $\mat{I}^{i}_{rgb}$ of object label $\boldsymbol{o}_j$, we propose a simple heuristic algorithm based on view frustum filtering and greedy matching. Firstly, the algorithm takes a list of 2D bounding boxes for each camera view, as well as camera intrinsic and extrinsic parameters, as input. It checks which points are in the 2D bounding box with a frustum filter. Then, a 2D bounding box $\vec{b}^{ij}_{2D}$ can be represented by the binary vector $\vec{v}^{ij}_{2D}\in \{0,1\}^N$, indicating the projection of which points out of all $N$ points are within the bounding box. We use the one-dimension Intersection Over Union (IOU) of two binary vectors as the matching score $s$ between two bounding boxes: 
\begin{equation}
\begin{aligned}
    s(\vec{b}^{ij}_{2D}, \vec{b}^{kl}_{2D}) &= \text{IOU}(\vec{v}^{ij}_{2D}, \vec{v}^{kl}_{2D})  \\
    &= \frac{\sum_{n=1}^{N} \min(\vec{v}^{ij}_{2D}[n], \vec{v}^{kl}_{2D}[n])}{\sum_{n=1}^{N} \max(\vec{v}^{ij}_{2D}[n], \vec{v}^{kl}_{2D}[n])},
\end{aligned}
\end{equation}
where $\vec{v}^{ij}_{2D}$ and $\vec{v}^{kl}_{2D}$ represent the binary vectors of the bounding boxes $\vec{b}^{ij}_{2D}, \vec{b}^{kl}_{2D}$ from views $i$ and $k$. This heuristic is predicated on the premise that the greater the number of points shared by two bounding boxes across views, the higher the likelihood of these bounding boxes pertaining to the same object. An demonstration of cross-view bounding box matching score computation is shown in Fig. \ref{fig:cross_view}.

With cross-view pairwise matching scores, the problem can be formulated as a quadratic assignment problem (QAP), which is known to be an NP-hard problem~\cite{jia2016roml}. We use a greedy algorithm approach to match bounding boxes across different views by maximizing the Intersection over Union (IoU) of these one-hot vectors, subject to the constraint that each bounding box can be matched at most once. The matched bounding boxes are then output as a list of tuples, with each tuple containing the indices of matched bounding boxes across views or "$-1$" if no match is found. It's important to note that there could be label discrepancies or failures across different camera views. We unify the object label of all matched 2D bounding boxes by max voting.

\subsubsection{\textbf{3D Instance Segmentation}}
With matched 2D bounding boxes for object $\mat{B}^{*j}_{2D}=\{\vec{b}^{ij}_{2D}\},\ i\in\{0,\cdots,k-1\}$, obtaining the objects' 3D instance segmentations is straightforward. Simply filter the point cloud within the matched bounding boxes across different views, and we can get the 3D instance segmentation of the object. Note that the shape of $\mat{B}^{*j}_{2D}$ is not fixed since there could be detection failures in random camera views.

In this section, we have delved into the workings of the RoboScript pipeline from receiving task instruction to execution, and explored its key components, code generation and perception. Since the Motion Planning is mainly done by Moveit, we do not repeat it here due to space constraints and will turn our attention to the discussion of our benchmark from the perspective of language model evaluation.

%% file: sec/4_benchmark.tex
\section{RoboScript Benchmark}
\label{sec:roboscript-benchmark}
Built on top of the infrastructure of Gazebo, MoveIt, and ROS, our pipeline generates executable code both in simulation and real robots, which enables us to evaluate the capability of language models to generate realistic deployable code, as well as to study the limitations and strengths of different language models in a real robotics system. For this purpose, we present our RoboScript benchmark. Compared with other benchmarks or test environments for general robot manipulation with natural language instructions \cite{zeng2021transporter,liang2023code,jiang2022vima}, our benchmark highlights realistic object manipulation, complex spatial reasoning with articulated objects, and a smaller gap with the real robotic system.

% Built on top of the infrastructure of Gazebo, MoveIt, and ROS, our pipeline generates executable code both in simulation and real robots. This feature is helpful in live systems since you can validate the correctness and safety of the control code before the robot executes the code. Further, this feature allows us to test the capability of language models to generate realistic deployable code, as well as to study the limitations and strengths of different language models in a real robotics system.

% For this purpose, we present our RoboScript benchmark. Compared with other benchmarks or test environments for general robot manipulation with natural language instructions \cite{zeng2021transporter,liang2023code,jiang2022vima}, our benchmark highlights realistic object manipulation, complex spatial reasoning with articulated objects, and a smaller gap with the real robotic system. \textcolor{red}{Junting: How can we better state the motivation for this benchmark???} Within this section, we will first discuss the general benchmark setup in \subsectionautorefname~\ref{subsec:benchmark-environment}. Following that, we will introduce the free-form tasks in \subsectionautorefname~\ref{subsec:free-form tasks}. Lastly, we will delve into the design of perception and motion planning tools in \ref{subsec:percetion_tools}.

\subsection{Environment Setup}
\label{subsec:benchmark-environment}

\begin{figure}
    \centering
    \includegraphics[width=1\linewidth, trim={1.5cm, 1.5cm, 1.5cm, 1.5cm}, clip]{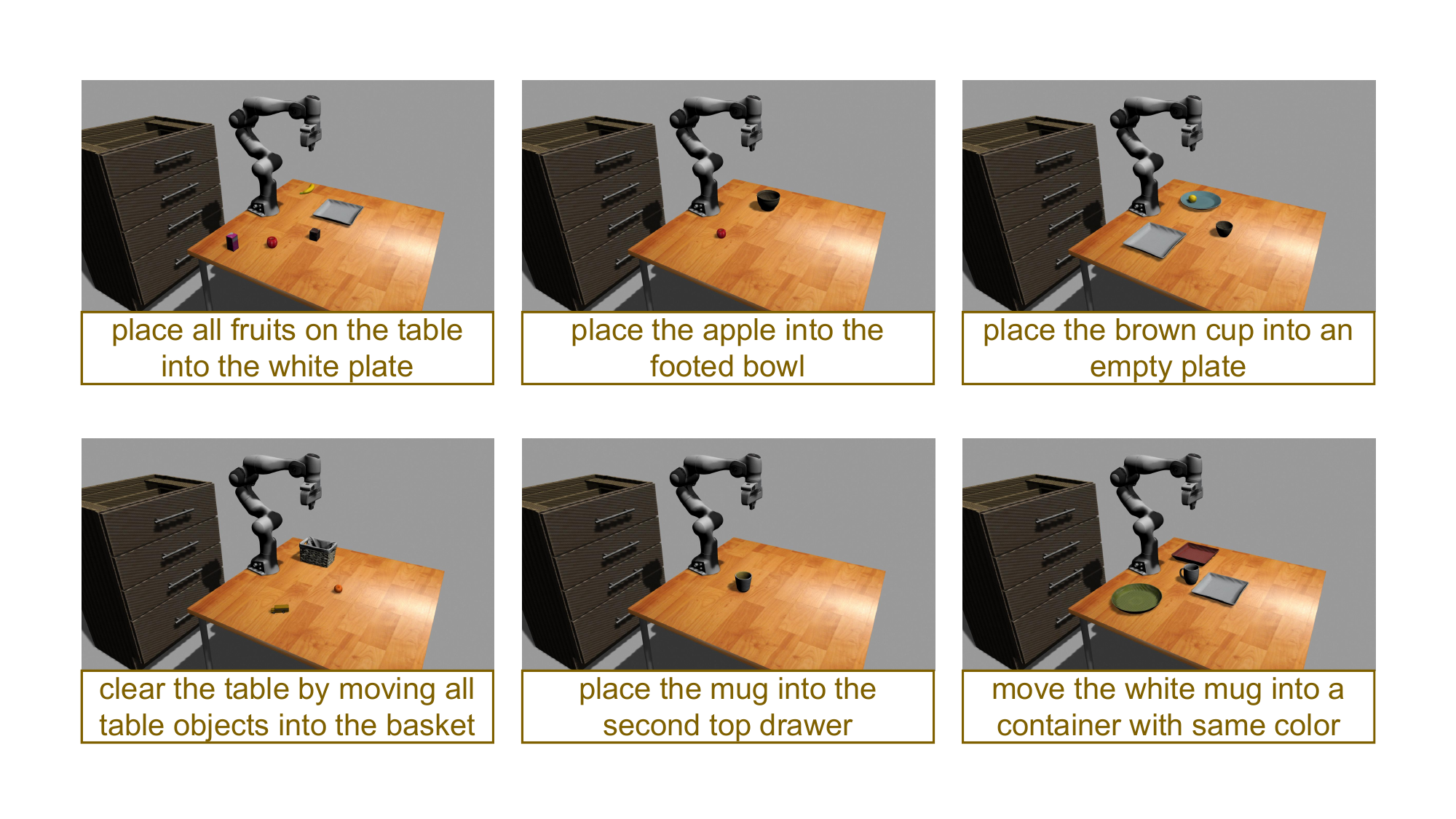}
    \caption{\textbf{Benchmark tabletop environment and tasks.}}
    \label{fig:worlds_and_tasks}
    \vspace{-15pt}
\end{figure}

The foundation of the RoboScript benchmark lies in its carefully designed base environment, which includes the following key elements:

\begin{itemize}
  \item \textbf{Robot Arm}: Our platform supports testing for 2 different robot arms: the Franka arm and the UR5 robot arm. The Franka arm is a lightweight and flexible 7-axis robot designed for research and manufacturing automation\cite{haddadin2022franka}. The UR5 is a versatile 6-axis collaborative industrial robot known for its flexibility, precision, and safety \cite{kebria2016kinematic}.
  % \item \textbf{Robot Arm}: Our platform supports testing for 3 different robot arms: the Franka arm, UR5, and Flexiv robot arm. The Franka arm is a lightweight and flexible 7-axis robot designed for research and manufacturing automation. It features advanced torque sensors and safety features for close human-robot collaboration \cite{haddadin2022franka}. The UR5 is a versatile 6-axis collaborative industrial robot known for its flexibility, precision and safety. It can be customized for various applications from assembly to machine tending \cite{kebria2016kinematic}. The Flexiv robot arm is a  6-axis articulated robot developed for factory automation and manufacturing. It provides good precision and path repeatability at an affordable price point. 
  
  \item \textbf{Furniture and Objects}:
  To simulate real-world scenarios, we have incorporated 10 cabinets with drawers and doors as parts from the \textit{storageFurniture} category in the PartNet-Mobility dataset ~\cite{xiang2020sapien}. Both the drawers and doors are articulated objects that require a deep understanding of physical constraints to manipulate successfully.
  % To simulate real-world scenarios, we have incorporated 10 cabinet with drawers and doors as part from storageFurniture category from PartNet-Mobility dataset~\cite{xiang2020sapien}. Both the drawers and doors are articulated objects and needs deeply understanding of the physical constrains to manipulate successfully.
  
  % such as a cabinet with multiple drawers and a table. These pieces of furniture serve as interaction points for the robot. drawers from 5 typical cabinets taken from PartNet-Mobility dataset~\cite{xiang2020sapien}
  
  \item \textbf{Diverse Containers}: For object interaction and manipulation tasks, we populate the environment with diverse containers. These containers are collected from the Google Scanned Objects 1k dataset~\cite{downs2022google}, representing some of the largest indoor object meshes sourced from real scans. The inclusion of these containers adds a layer of realism and complexity to the tasks.
  
  \item \textbf{Pickable Objects}: The benchmark includes 57 pickable objects, carefully selected from the YCB dataset~\cite{calli2015benchmarking}. This dataset encompasses a wide array of objects, including both household and kitchen items, ensuring a diverse set of objects for the robot to interact with.
\end{itemize}

\subsection{Free-form Manipulation Tasks}
\label{subsec:free-form tasks}

As shown in \tableautorefname~\ref{tab: tasks_definition} and Fig. \ref{fig:worlds_and_tasks}, our benchmark comprises eight carefully designed tasks featuring four distinct challenges that test the reasoning capabilities of LLMs. The manipulation tasks outlined in the benchmark table are cleverly designed to assess various levels of cognitive processing in language models. The simplest task involves placing an apple into a bowl, testing basic category understanding, object selection, and spatial awareness—suitable for an `Easy' difficulty level. As the tasks become more nuanced, we see `Moderate' challenges like nesting objects (a mug into a drawer) or categorizing and repositioning items (fruits onto a plate, or swapping an apple and a banana), which require a blend of categories and spatial reasoning. A task such as placing a brown cup onto an empty plate introduces property reasoning, adding a layer of complexity. The `Hard' tasks are the most intricate: clearing a table by moving all objects into a basket demands advanced spatial and property reasoning while placing a white mug into a container of the same color tests the model's ability to comprehend and apply abstract properties like color coordination. These tasks are thoughtfully curated to evaluate the model's understanding from simple object recognition to complex, multi-step reasoning.

\begin{table*}[!htp]\centering
\caption{\textbf{Free-form manipulation tasks in the RoboScript benchmark.} The benchmark comprises eight carefully designed tasks of three difficulty levels, featuring four distinct challenges that test the reasoning capabilities of LLMs.}
\label{tab: tasks_definition}
\vspace{2pt}
\scriptsize
\resizebox{0.95\linewidth}{!}{%
\begin{tabular}{lccccr}\toprule
\textbf{Task ID} &\multicolumn{4}{c}{\textbf{Difficulties in Language Model Reasoning}} &\textbf{Task Instruction} \\\cmidrule{1-6}
\textbf{} &\textbf{Category Understanding} &\textbf{Object Selection} &\textbf{Spatial Relationship} &\textbf{Property Reasoning} &\textbf{} \\\midrule
Simple\_0 & & & & &Place the apple into the footed bowl \\
Simple\_1 & & &\checkmark & &Place the brown cup into an empty plate \\
Moderate\_0 &\checkmark &\checkmark &\checkmark & &Open the second top cabinet drawer and place all boxes into it \\
Moderate\_1 & & & &\checkmark &Move the white mug into a container with the same color \\
Moderate\_2 & & &\checkmark & &Place the mug into the second top drawer \\
Hard\_0 &\checkmark &\checkmark & & &Place all fruits on the table into the white plate. \\

Hard\_1 & & &\checkmark & &Exchange the position of the apple and the banana on the table \\

Hard\_2 & &\checkmark & & &Clear the table by moving all table objects into the basket \\

\bottomrule
\end{tabular}}
\vspace{-10pt}
\end{table*}

% \subsection{Large Multi-Modal Models and Large Language Models}

\subsection{Perception and Motion Planning Tools}
\label{subsec:percetion_tools}
\subsubsection{Grasp Pose Estimation}
We provide 4 types of grasp pose estimation methods. 
\textbf{a.~Central Lift Grasp:} As a heuristic-based method provided in the HomeRobot project~\cite{homerobotovmm}, the system executes a top-down grasp by aligning the gripper directly above the object's geometric center. 
\textbf{b.~Horizontal Grasping:} In this approach, the gripper is oriented horizontally, making it suitable for objects aligned with a plane perpendicular to the tabletop. 
\textbf{c.~GIGA Grasp Pose Prediction~\cite{jiang2021synergies}:} 
GIGA presents a neural network-based approach for 6-DOF grasp pose prediction. This method utilizes generative models and offers enhanced flexibility. However, it may have limitations in constrained tasks.
\textbf{d.~AnyGrasp~\cite{fang2023anygrasp} Grasp Pose Prediction:} AnyGrasp, a state-of-art neural network-based grasping model, is adept at generating dense 6-DOF grasp poses. It is trained on the comprehensive GraspNet dataset \cite{fang2020graspnet}, which encompasses over 1 billion grasp labels, ensuring a robust and versatile grasping capability.

% \subsubsection{Grasp Pose Estimation}
% We provide 4 types of grasp pose estimation methods. \textbf{Central Lift Grasp:} As a heuristic based method provided in the HomeRobot project\cite{homerobotovmm}, it grasps objects directly from above the central point with a top-down pose. \textbf{Horizontal Grasping:} In this approach, the gripper is oriented horizontally, making it suitable for objects aligned with a plane perpendicular to the tabletop. \textbf{AnyGrasp~\cite{fang2023anygrasp} Pose Prediction:} 
%   AnyGrasp is a neural networks grasping model and is trained on the GraspNet dataset\cite{fang2020graspnet} which contains 1B grasp labels.
%  \textbf{GIGA Pose Prediction~\cite{jiang2021synergies}:} Another neural network-based approach, GIGA, is employed for grasp prediction. This method utilizes generative models and offers enhanced flexibility. However, it may have limitations in constrained tasks. 
  % The choice between AnyGrasp and GIGA depends on the specific requirements of the task, the objects involved, and the environment.

\subsubsection{Kinematic Modeling of Articulated Objects}
We utilize the GAMMA\cite{yu2023gamma} to predict articulated objects' kinematic structures. 
GAMMA effectively processes the object's point clouds, segments them into rigid parts, and accurately estimates the associated kinematic parameters. More specifically, GAMMA predicts offsets for each point towards part centroids, thereby grouping them into distinct articulated components. Additionally, the model conducts regression on per-point projections onto joint axes, facilitating a voting process for determining axis origins. Simultaneously, it regresses per-point joint directions, aggregating votes to establish precise axis directions. GAMMA demonstrates the capability to generically model both revolute and prismatic joint parameters by clustering adjusted and projected points, entirely independent of the articulated object's category.

\subsubsection{Motion Planning Module}
% The motion planning module takes as input the current pose and goal pose provided by the prediction module, and outputs corresponding waypoints trajectories. 
Our planning module is built upon the Open Motion Planning Library (OMPL) platform~\cite{sucan2012open} integrated within the MoveIt platform, which enables performance benchmarking across 12 different motion planning algorithms including RRT* \cite{karaman2011sampling}, PRM*~\cite{gang2016prm} and so on. For this work, we primarily utilized the RRT* algorithm due to its efficient exploration and optimality guarantees. The planning module samples configuration space to find a collision-free path to connect the start and goal states. The output trajectory is checked for dynamic feasibility before being passed to the control module for execution.

\begin{figure}[h]
\vspace{-5pt}
    \centering
    \begin{subfigure}[b]{0.23\textwidth}
        \includegraphics[height=3.5cm]{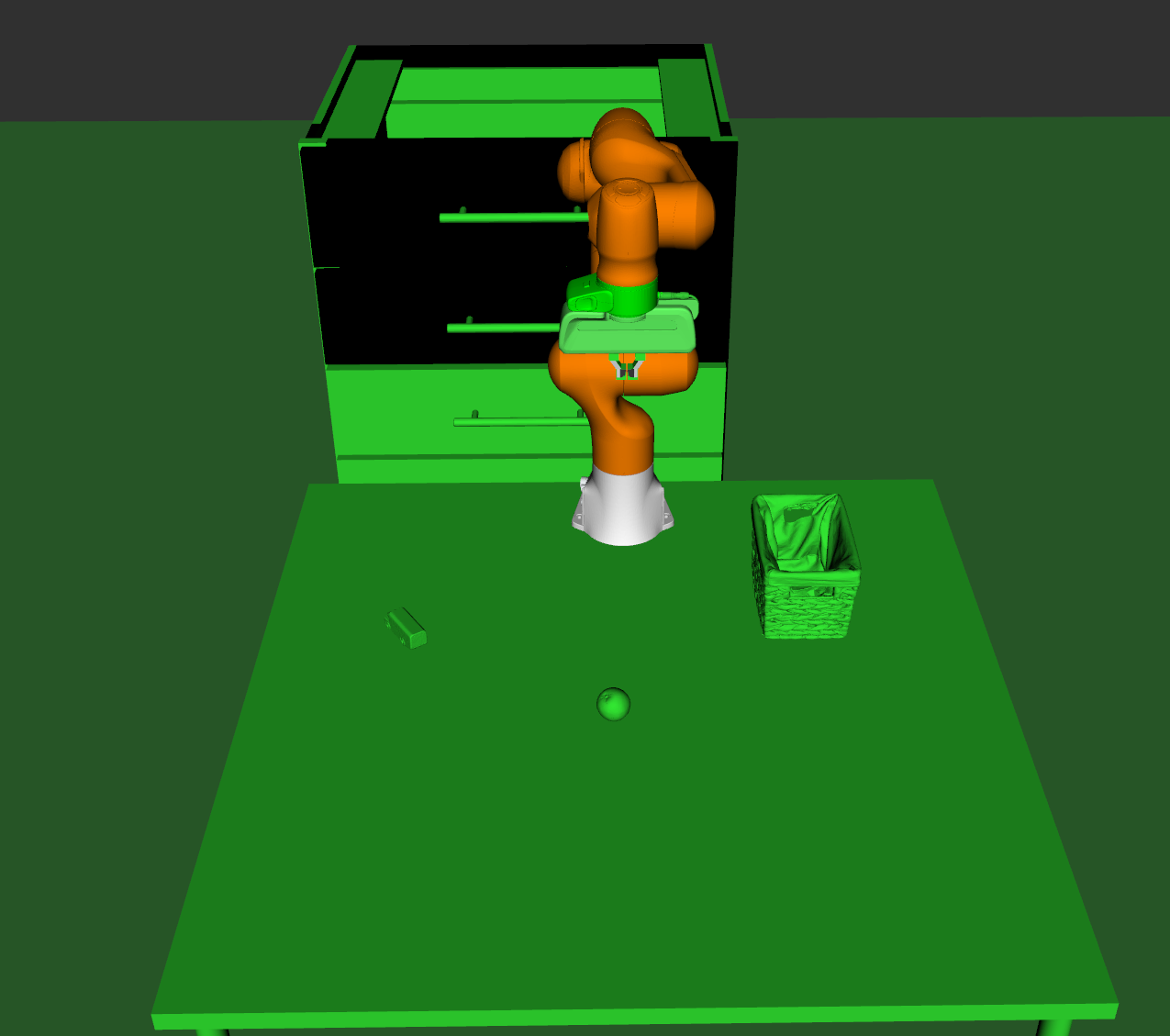}
        \caption{Run MoveIt with GT mesh}
        \label{fig:image1}
    \end{subfigure}
    % \hfill
    \begin{subfigure}[b]{0.23\textwidth}
        \includegraphics[height=3.5cm]{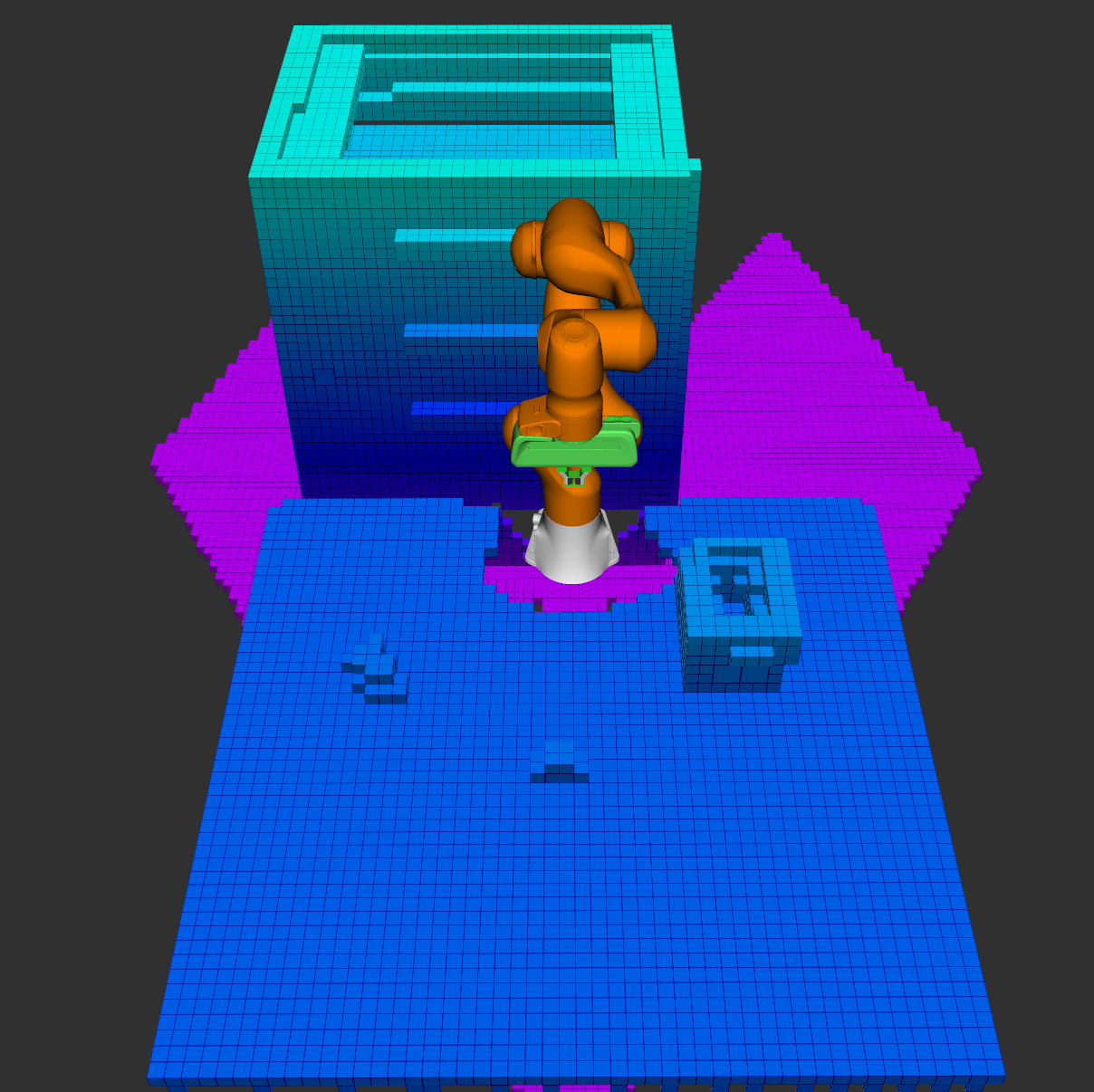}
        \caption{Run MoveIt with sensor input}
        \label{fig:image2}
    \end{subfigure}
    \vspace{2pt}
    \caption{\textbf{MoveIt planning scene.} There are two modes for running MoveIt planning scene in the benchmark. MoveIt can either load ground truth mesh data from Gazebo, or build an Octomap from sensors in real time and load reconstructed meshes from the perception pipeline.
planning scene}
    \vspace{-8pt}
    \label{fig:moveit_vis}
\end{figure}

In the benchmark, we offer two modes for running the MoveIt planning scene, which is the data structure used to store objects and obstacles for collision checking during planning. Users can either 1) enable using ground truth planning scenes, with perfect model meshes of Gazebo world loaded into the MoveIt planning scene and updated automatically; or 2) enable the planning scene sensor plugin to construct and update an Octomap from depth images in the real-time. The meshes of objects to be contacted or taken as receptacles are loaded at request by LLMs when calling ``\textit{detect\_objects}'' function. The visualization of MoveIt planning scene under the two options is shown in Fig.~\ref{fig:moveit_vis}.

% To improve planning efficiency, we apply a smoothing technique to shorten the generated RRT* paths.  
% We validated the performance of our planning module in both simulation and real-world experiments, demonstrating reliable trajectory generation for navigation in cluttered environments. 

% We first discuss our design choices for implementing~\algname below. Then we validate directly in real-world systems whether~\algname can perform everyday manipulation tasks in Sec.~\ref{sec:real}. We also present a detailed quantitative study of the generalization performance of~\algname compared to learned and LLM-based baselines in simulation in Sec.~\ref{sec:sim}. We further demonstrate how~\algname can benefit from only limited online experience to learn a dynamics model for contact-rich tasks in Sec.~\ref{sec:dynamics-exp}. Finally, we study the source of errors in the overall systems and discuss how improvement can be made in Sec.~\ref{sec:error}.
% We first discuss our implementation details. Then we validate~\algname{\alias} for real-world everyday manipulation (Sec.~\ref{sec:real}). We also study its generalization in simulation (Sec.~\ref{sec:sim}). We further demonstrate how~\algname{\alias} enables efficient learning of more challenging tasks (Sec.~\ref{sec:dynamics-exp}). Finally, we analyze its source of errors and discuss how improvement can be made (Sec.~\ref{sec:error}). 

%% file: sec/5_experiments.tex
\section{Experiments}
\label{sec:exp}

In this section, we first present the LLMs evaluation results on the RoboScript pipeline, discussing the strengths and limitations of LLMs for realistic control code generation in \subsectionautorefname~\ref{exp: llm_eval}. We then conduct an in-depth study of how the physics properties of objects influence grasp and arm motion, and subsequently, the final success rate of tasks in \subsectionautorefname~\ref{exp: object_shape}. Lastly, in~\subsectionautorefname~\ref{exp: perception}, we present an ablation study of the perception pipeline, demonstrating how perception could potentially bottleneck the entire model. We also present our findings on how language models and vision models collaborate can make the pipeline work efficiently, and what happens otherwise.

\vspace{-5pt}
\subsection{Benchmarking LLMs on RoboScript}
\label{exp: llm_eval}
In this subsection, we present the evaluation results of three popular LLMs on the RoboScript benchmark. Table \ref{tab:llm_eval} shows the evaluation result under the ground truth perception settings, where the perception pipeline is turned off and the ground truth 3D perception from the Gazebo simulation is always used. The results are the average over 10 repeated trials. There are four metrics presented in the results: 1) \textit{Grammar correctness}, which indicates whether the generated script has correct grammar; 2) \textit{Semantic correctness}, which indicates whether the generated script can faithfully complete the task as per the language instruction without logical errors. This is determined by the success of any of the 10 trials. However, we also manually check each code snippet that fails all trials, to eliminate randomness in the reported results; 3) \textit{Finished steps ratio}, which indicates the average success rate of all pre-defined steps of a task. For example, task hard\_0 involves picking two fruits on a plate, which comprises two steps: moving the first fruit and moving the second fruit; 4) \textit{Finished whole task}, which indicates the success rate of the entire task. We further collect all failed snippets in Appendix~\ref{appendix: failed_code} and analyze the failure reasons. 

From the perspective of evaluating the code generation ability of LLMs, the first two metrics are more meaningful, as the task execution success rate depends more on other factors, including grasp pose detection, motion planning, etc. From Table \ref{tab:llm_eval}, we can see that GPT-4 outperforms the other two LLMs by a clear margin. GPT-4 reaches amazingly $100\%$ correctness on both grammar and semantics. Surprisingly, despite our pre-defined prompt introduced in section \ref{pipeline: code_generation}, Gemini-pro performs even worse than GPT-3.5-turbo, in terms of code generation quality. This result was observed even after we provided Gemini-pro with an additional instruction prompt:
\vspace{-5pt}
\begin{framed}
\vspace{-2pt}
\small{\textit{Please pay attention to the format of examples in chat history and imitate it. Do NOT generate common texts or code to import any Python libraries. ONLY generate comments or detailed planning starting with \# and Python code.}}
\vspace{-3pt}
\end{framed}
Without this prompt, Gemini-pro fails almost all tasks due to the random output format. Through the case study, we found that Gemini-pro exhibits more hallucinations compared to GPT-3.5 or GPT-4, and pays less attention to instructions in the system prompt. In the Gemini technical report \cite{team2023gemini}, there is no mention of techniques similar to system message tokens as used in Llama-2 and GPTs. This could be related to the issue, but no definitive conclusion can be drawn.

\begin{table*}[!htp]\centering
\caption{\textbf{Evaluation result of LLMs on the RoboScript benchmark.} There are four metrics presented in the results: 1) \textit{grammar correctness}, which indicates whether the generated script has correct grammar; 2) \textit{semantic correctness}, which indicates whether the generated script can faithfully complete the task as per the language instruction without logical errors. 3) \textit{finished steps ratio}, which indicates the average success rate of all  steps of a task. 4) \textit{finished whole task}, which indicates the success rate of the entire task.}
\label{tab:llm_eval}
\scriptsize
\begin{adjustbox}{width=0.9\linewidth,center}
\begin{tabular}{llccccccccc}
\toprule
 LLMs & Metric & Simple\_0 & Simple\_1 & Moderate\_0 & Moderate\_1 & Moderate\_2 & Hard\_0 & hard\_1 & Hard\_2\\
\midrule
\multirow{2}{*}{ Gemini-pro } & grammer correctness & false & false & false & false & true & true & false & true \\
& semantic correctness & false & false & false & false & true & false & false & false \\
& finished steps ratio & 0 & 0 & 0 & 0 & 0.2 & 0.2 & 0 & 0.3 \\
& finished whole task & 0 & 0 & 0 & 0 & 0.2 & 0 & 0 & 0 \\
\hline \multirow{4}{*}{ GPT-3.5-turbo } & grammer correctness & true & false & true & true & true & true & true & false \\
& semantic correctness & true & false & true & true & true & false & false & false \\
& finished steps ratio & 0.8 & 0 & 0.3 & 0.35 & 0.7 & 0 & 0 & 0 \\
& finished whole task & 0.8 & 0 & 0.3 & 0.1 & 0.7 & 0 & 0 & 0 \\
\hline
\multirow{4}{*}{ GPT-4 } & grammer correctness & true & true & true & true & true & true & true & true \\
& semantic correctness & true & true & true & true & true & true & true & true \\
& finished steps ratio & 0.7 & 0.9 & 0.1 & 0.65 & 0.4 & 0.55 & 0.15 & 0.45 \\
& finished whole task & 0.7 & 0.9 & 0.1 & 0.4 & 0.4 & 0.3 & 0 & 0 \\
\toprule
\end{tabular}
\end{adjustbox}
\vspace{-10pt}
\end{table*}

% In this section, we present and analyze the results of code generation by different large language models on our benchmark. However, the task coverage is limited by a pre-defined world layout and choices of objects. In the next section, we will further study the impact of objects instead of tasks.

\subsection{Grasp and Motion Plan: Impact of Object Shapes}
\label{exp: object_shape}
In this section, we shift our focus from the code generation and reasoning capabilities of LLMs to lower-level factors: the environment objects and how their shapes affect the manipulation task. Through the construction of the benchmarking world and the selection of appropriate objects for different tasks, we find that object shapes dramatically impact the success of object manipulation for a gripper with certain mechanical structures. To better study this impact, we isolate the errors in LLM reasoning and code generation by selecting two fundamental manipulation tasks in daily scenarios: 1) object pick and place on a table (Simple\_0); 2) placing objects into a closed cabinet drawer (Moderate\_1). We execute verified scripts based on code generated by GPT-4 and only change the drawers and objects. 

\begin{figure}
\vspace{-12pt}
    \centering
    \begin{subfigure}[b]{0.93\linewidth}
        \includegraphics[width=\textwidth]{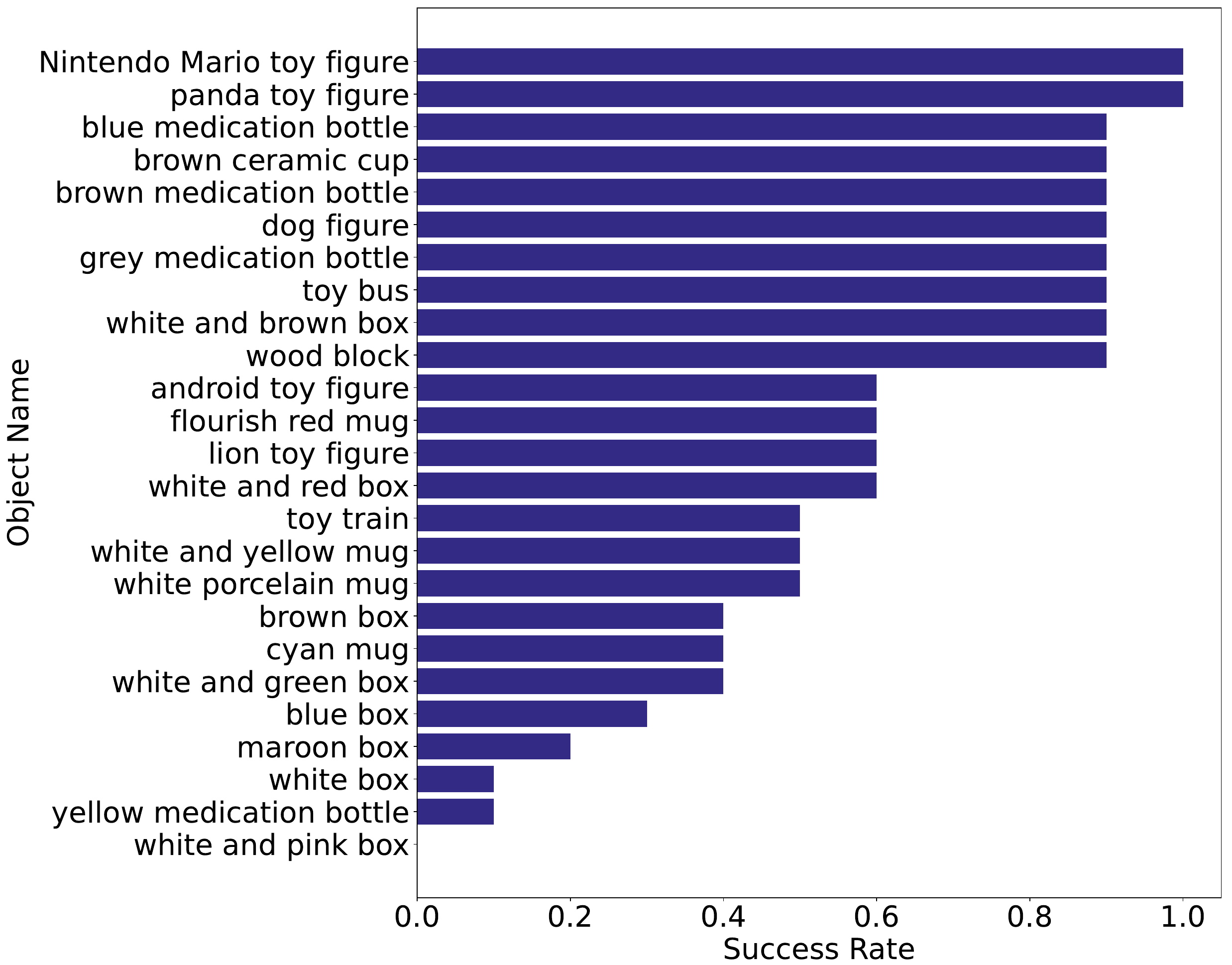}
        % \vspace{-10pt}
        \caption{Google Scanned Objects}
        % \label{fig:subfig1}
    \end{subfigure}
    \hfill
    \begin{subfigure}[b]{0.93\linewidth}
        \includegraphics[width=\textwidth]{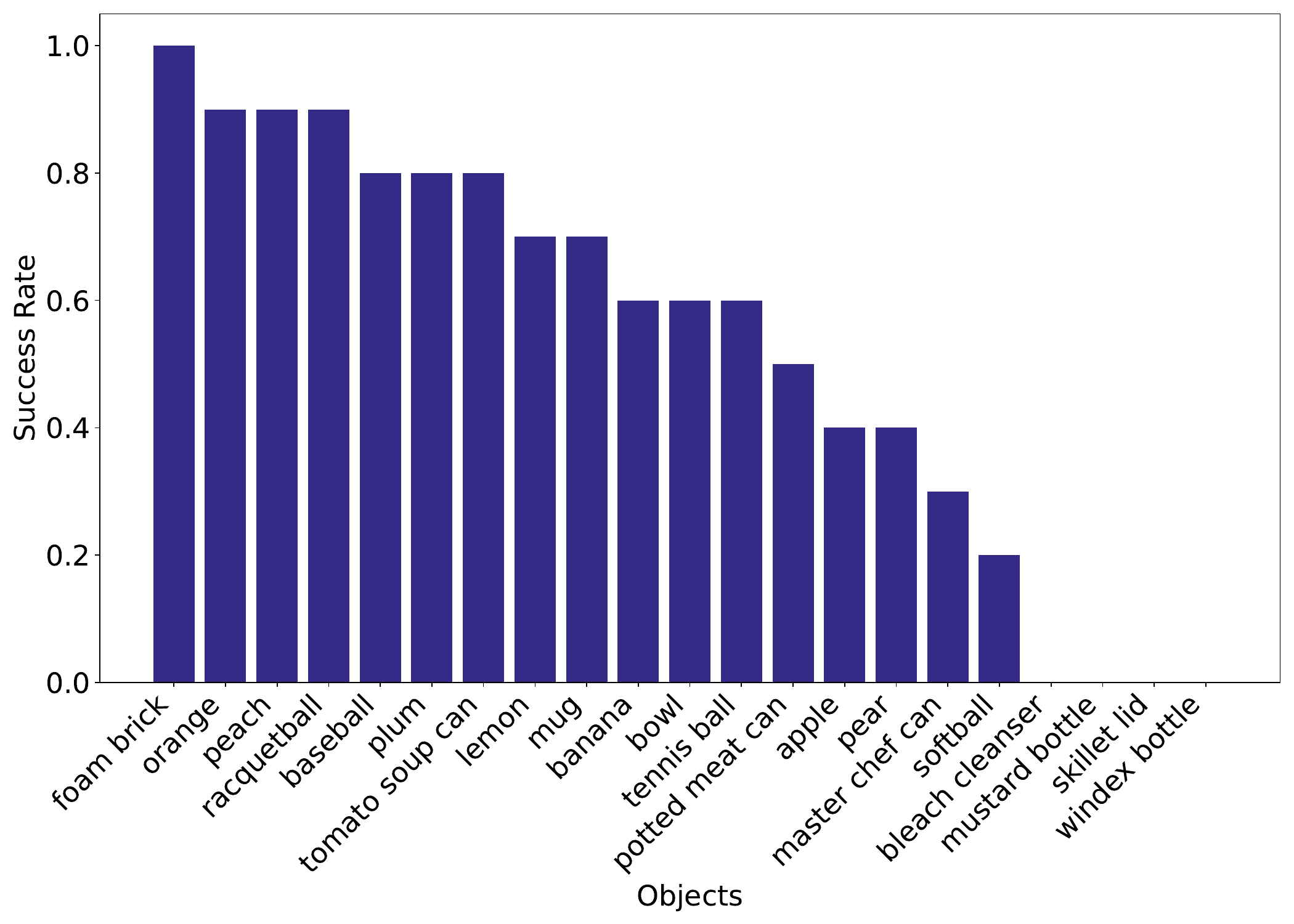}
        \caption{YCB objects}
        % \label{fig:subfig2}
    \end{subfigure}
    \caption{\textbf{Results of pick and place task.} They present the success rates of pick and place tasks for all object models in our model base, including 21 YCB objects and 25 Google Scanned objects.}
    \vspace{-17pt}
    \label{fig:pick_and_place}
\end{figure}

First, we exhibit the results of table object pick and place in Fig. \ref{fig:pick_and_place}. Among our object models, 21 come from YCB \cite{calli2015ycb} and 25 come from Google Scanned Objects \cite{downs2022google}.

Secondly, we select the objects with the top success rates in tabletop pick and place task (Fig.~\ref{fig:pick_and_place}), and test their success rates in "open a drawer, pick and place an object into the drawer" task. We present results in Table~\ref{tab:drawer_pnp_google} and \ref{tab:drawer_pnp_ycb}.

A striking observation is that some round and cylindrical objects, like the \textit{brown\_medication\_bottle} and \textit{grey\_medication\_bottle} from Google Scanned Objects and \textit{orange} and \textit{baseball} from YCB dataset perform well in desktop pick and place task but struggle when placed in the drawer. Through careful investigation, we found that when using a thin two-fingered mechanical gripper to grasp cylindrical and spherical objects, the grasp is always in an unstable mechanical state. The end of the gripper could not always maintain a Cartesian path when moving in a complex constrained space. And sudden accelerations and rotations would cause this mechanical state to go out of control.

\begin{table}[!htp]\centering
\caption{\textbf{Success rate of task opening drawer\_0 and placing Google Scanned Objects.} We observe two cylindrical objects have significant performance drop compared to pick and place task.}\label{tab:drawer_pnp_google}
\scriptsize
\resizebox{0.8\linewidth}{!}{%
\begin{tabular}{lrrr}\toprule
\textbf{object} &\textbf{open\_drawer} &\textbf{place\_object}  \\
\midrule
wood\_block &1 &0.8 \\
\rowcolor{gray!20} grey\_medication\_bottle &0.9 &0.4 \\
white\_and\_brown\_box &1 &0.6 \\
panda\_toy\_figure &1 &0.9 \\
\rowcolor{gray!20} brown\_medication\_bottle &1 &0.2 \\
blue\_medication\_bottle &0.9 &0.7 \\
brown\_ceramic\_cup &0.9 &0.5 \\
toy\_bus &1 &0.8 \\
Nintendo\_Mario\_toy\_figure &0.9 &0.8 \\
dog\_figure &1 &0.5 \\
\bottomrule
\end{tabular}
}
\vspace{-8pt}
\end{table}

\begin{table}[!htp]\centering
\caption{\textbf{Success rate of task opening drawer\_1 and placing YCB objects.} We observe that two sphere objects have a significant performance drop compared to the pick and place task.}\label{tab:drawer_pnp_ycb}
\scriptsize
\resizebox{0.8\linewidth}{!}{%
\begin{tabular}{lrrr}\toprule
\textbf{object} &\textbf{open\_drawer} &\textbf{place\_object}  \\
\midrule

plum &0.9 &0.4 \\
tomato\_soup\_can &0.8 &0.6 \\
lemon &1 &0.3 \\
\rowcolor{gray!20} orange &1 &0 \\
mug &0.9 &0.7 \\
\rowcolor{gray!20} baseball &0.6 &0.1 \\
lemon &1 &0.3 \\
tennis\_ball &1 &0.4 \\
banana &1 &0.3 \\
\bottomrule
\end{tabular}
}
\vspace{-15pt}
\end{table}

% In this section, we put aside LLM and code generation and discuss how different objects impact the fundamental capabilities of the robot arm and gripper. However, the perception tools of our pipeline remain untouched. In the next section, we will inspect this final piece of our puzzle. 

\subsection{Perception Matters: Pipeline Ablation Study}
\label{exp: perception}
In this section, we present the results of the pipeline ablation study, mostly focusing on how different perception modules affect the overall instruction-following manipulation pipeline. The perception pipeline generates 3D object instances that can be utilized by perception or motion planning tools. Thus, we study how 1) 3D object grounding and 2) input data of MoveIt planning scene affect the pipeline. To minimize the influence of object shape and grasp detection on the experiments, we selected the 10 objects from Google Scan Objects that had the highest grasp success rate in the last experiment. For each object model, each task repeats for 10 times and we report the success rate out of these 10 trials.

\begin{figure}
    \vspace{-10pt}
    \centering
    \includegraphics[width=1\linewidth]{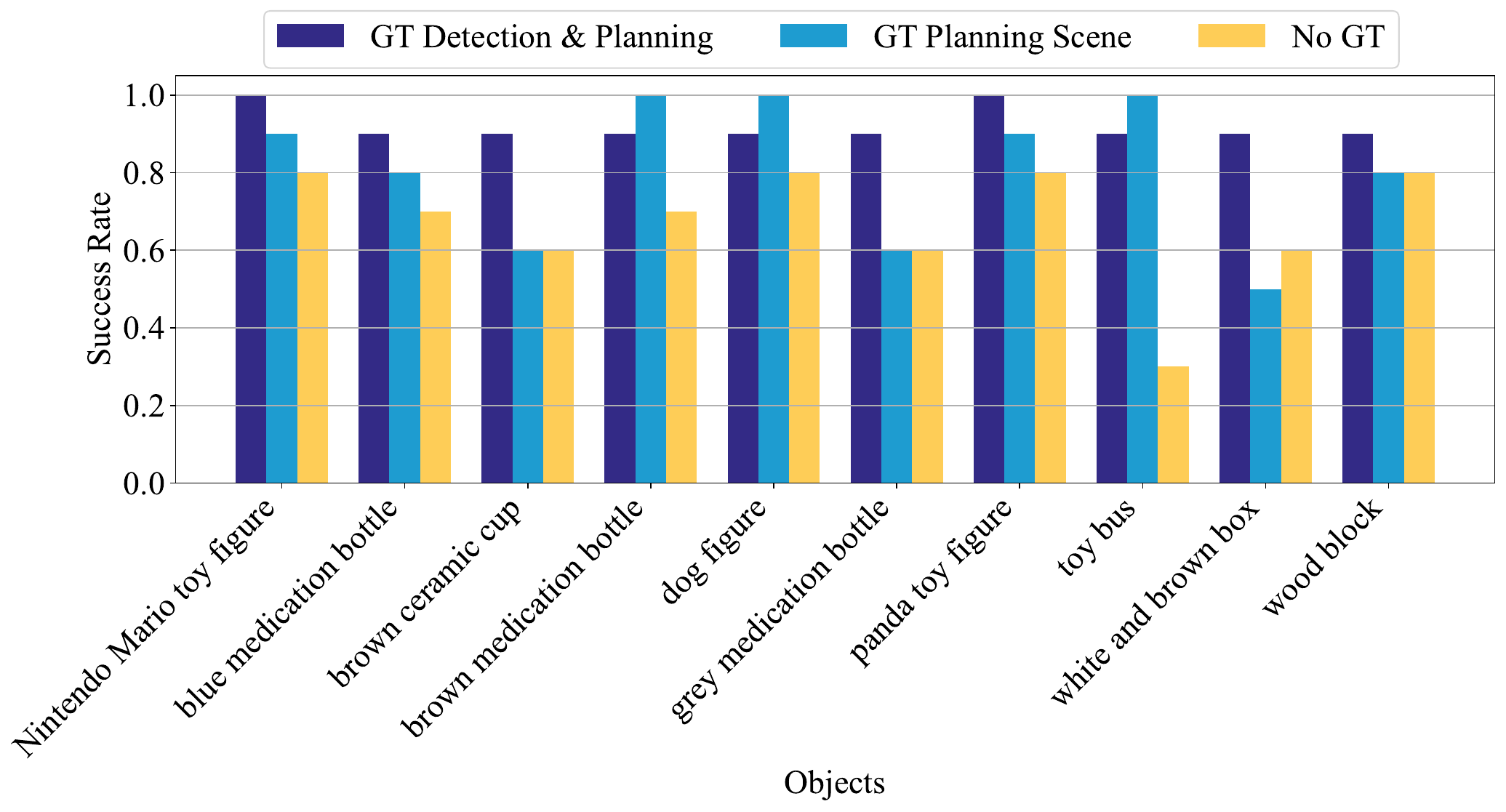}
    \caption{\textbf{Perception ablation study: pick and place.} The figure presents the success rate of task pick and place under three ablation settings: 1) GT Detection \& Planning, indicating the perception tools and motion planning tools use ground truth 3D representations from Gazebo; 2) GT Planning Scene, indicating the system only uses ground truth object meshes in motion planning tools from MoveIt; 3) No GT, indicating the system does not use ground truth data, all from perception pipeline.}
    \label{fig:ablation_pnp}
    \vspace{-8pt}
\end{figure}

\begin{figure}
    \centering
    \includegraphics[width=1\linewidth]{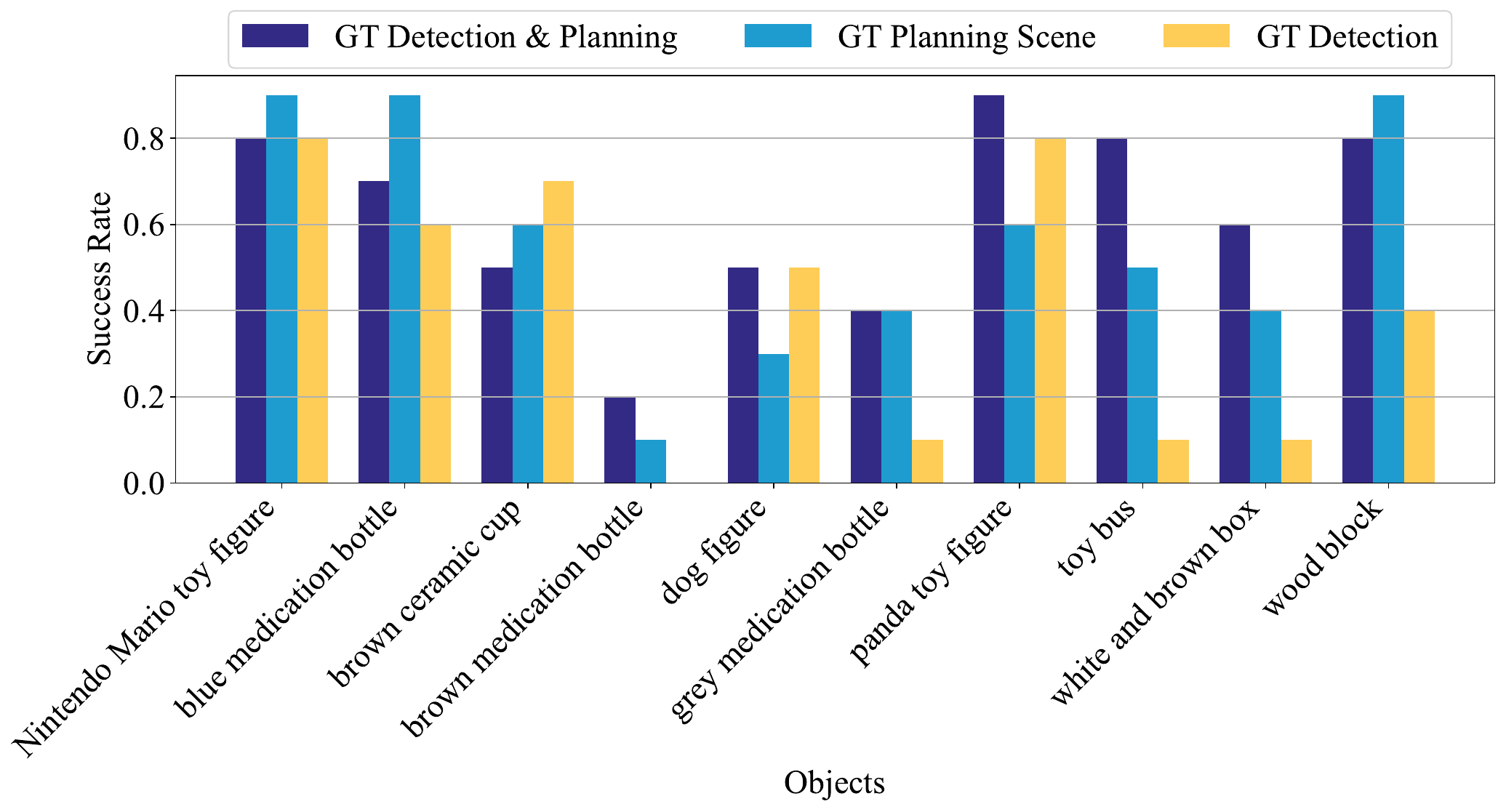}
    \caption{\textbf{Perception ablation study: open drawer and place objects.} The figure presents the success rate of task pick and place open the drawer and place objects under three ablation settings: 1) GT Detection \& Planning, indicating the perception tools and motion planning tools use ground truth 3D representations from Gazebo; 2) GT Planning Scene, indicating the system only uses ground truth object meshes in motion planning tools from MoveIt; 3) GT Detection, indicating the system only uses ground truth 3D object instances as the result of perception pipeline.}
    \label{fig:ablation_drawer_pnp}
    \vspace{-12pt}
\end{figure}

\begin{figure}
\vspace{-15pt}
    \centering
    \includegraphics[width=1\linewidth, trim={4cm, 2cm, 4cm, 2cm}, clip]{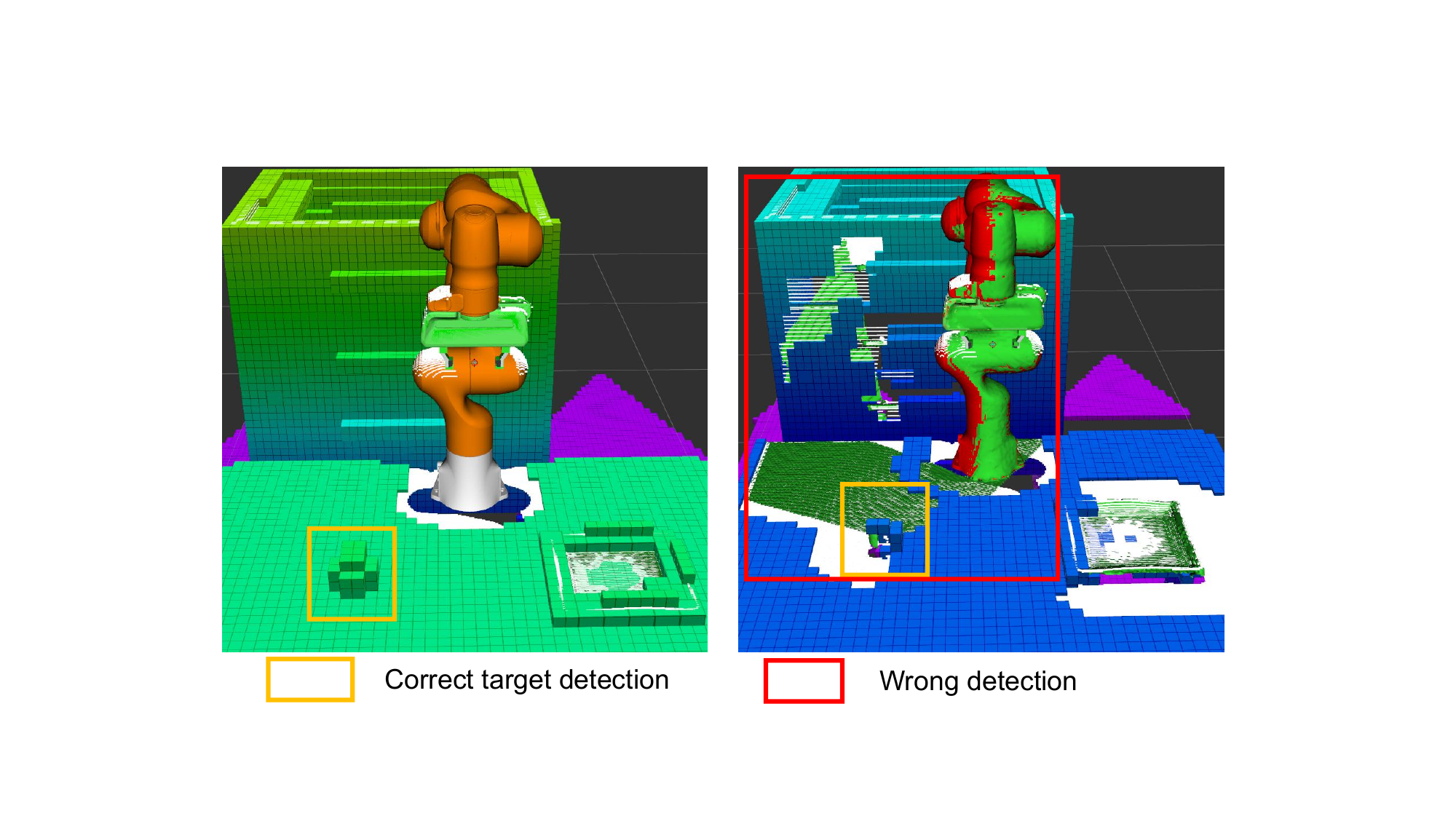}
    \caption{\textbf{Wrong perception leads to motion planning failure.} Incorrect 2D grounding can result in an incorrect 3D instance bounding box that encompasses the Franka robot. The robot arm's mesh can cause unsolvable self-collisions, rendering all states illegal for motion planning. As a result, the command execution is refused.}
    \label{fig:self_collision}
    \vspace{-10pt}
\end{figure}

Following the settings in last \subsectionautorefname, we test and report the system performance on 1) Pick and Place in Fig.~\ref{fig:ablation_pnp}; 2) Open Drawer and Place Object tasks in Fig.~\ref{fig:ablation_drawer_pnp}. By enabling the real perception pipeline, using 3D bounding boxes from 2D grounding and cross-view matching rather than ground truth in Gazebo, the success rate of pick and place dropped by $10.1\%$, and the success rate of task open drawer and place object dropped by $6.0\%$. In comparison, using 3D representations of the Octomap and reconstructed mesh from perception makes the performance drop by $14.0\%$ and $15.0\%$ on the two tasks separately. The result demonstrates that the noise in the perception pipeline has a larger impact on motion planning than perception tools, mostly the grasp pose detection models in our tasks. This could hint that the noise in Octomap construction is large, affecting the motion planning accuracy. However, we also find that MoveIt motion planning is sensitive to the error in detections, as demonstrated in Fig. \ref{fig:self_collision}. In this case, the bounding box of the target object is enlarged, resulting in an incorrect 3D instance bounding box that encompasses the Franka robot. The robot arm's mesh will cause unsolvable self-collisions, rendering all states illegal for motion planning. Thus, any motion request will be rejected.

\subsection{Real Robot Deployment}

\begin{figure}
    \centering
    \includegraphics[width=0.99\linewidth]{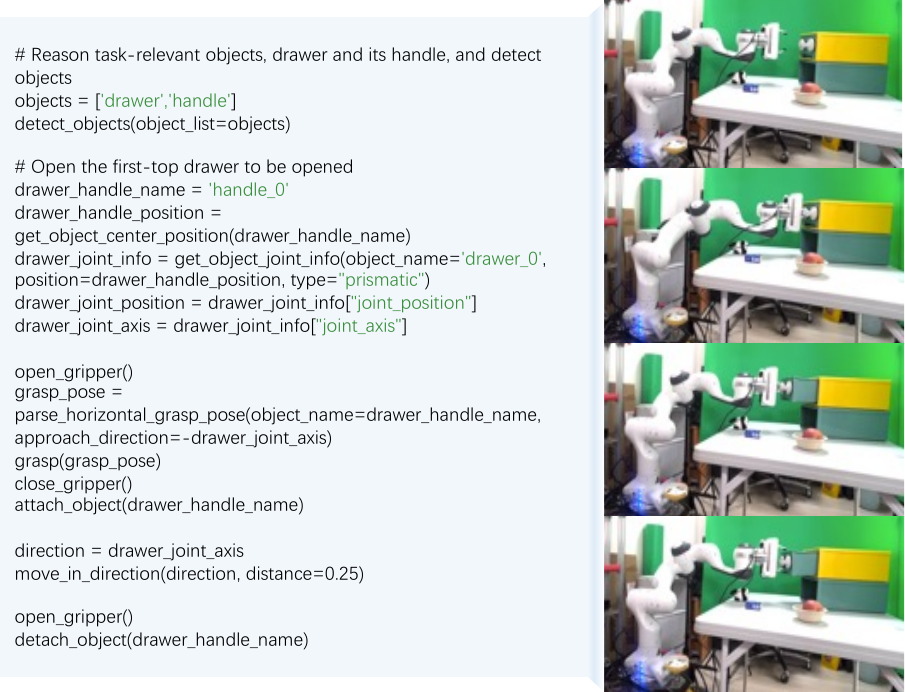}
    \caption{\textbf{Deployment on the real robot: open the drawer with the Franka robot.}}
    \label{fig:real_pnp}
    \vspace{-15pt}
\end{figure}

% To further support our claim that Roboscript is a unified code generation pipeline for both simulation testing and real robot deployment, we installed the RoboScript onto two sets of robot systems: Franka Panda and UR5 + Robotiq 85. 
To further support our claim that RoboScript is a unified code generation pipeline for both simulation testing and real robot deployment, we installed RoboScript onto two sets of robot systems: the Franka Panda and the UR5 with Robotiq 85 gripper. As shown in Fig. \ref{fig:real_pnp}, this deployment demonstrated RoboScript's ability to seamlessly integrate with existing robotic frameworks, ensuring that tasks such as object manipulation and precision placement are executed with high fidelity across different hardware configurations. Our findings indicate that RoboScript not only simplifies the transition from simulation to real-world application but also enhances the robots' adaptability to diverse tasks without requiring extensive programming effort. This dual deployment validates RoboScript's role as a pivotal technology in bridging the gap between simulation-based development and practical, real-world robotic automation.

% \zhixuan{Add the experiment result with and without CoT comments}

%% file: sec/6_conclusions.tex
\vspace{-5pt}
\section{Conclusion}
\label{sec:con}

This paper presents \alias that facilitates the deployment and evaluation of large language models for generating executable robot control code. By integrating with simulation and real-world robotics frameworks like ROS, Gazebo, and MoveIt, \alias bridges high-level reasoning and low-level control. The comprehensive pipeline encompasses perception, planning, grasp detection, and motion control achieving an end-to-end system from language interpretation to plan execution. Experiments highlight the GPT model's reasoning in complex spatial, physical, and property reasoning manipulation tasks. Ablation studies further illustrate the impact of factors like object geometry, perception accuracy, and planning robustness on real-world deployment. Overall, RoboScript advances the integration of language model intelligence with robots for autonomous manipulation. Future work can focus on enhancing open-world reliability, expanding the scope of instructions/environments, and improving sim-to-real transfer sample efficiency to further leverage the potential of AI-powered systems.

%% file: sec/7_appendix.tex
\section*{Appendix}
\label{sec:appendix}

\subsection{Code generation algorithm}
\begin{algorithm}[htb]
\footnotesize
\KwData{Main system\_prompt $S$, Main examples $E$, Sub system\_prompt $S^\prime$, Sub examples $E^\prime$, task\_query $Q$}
\KwResult{Generated code following the hierarchical agent structure}

\SetKwProg{Class}{Class}{}{}
\SetKwFunction{FMain}{MainAgent}
\SetKwFunction{FSub}{SubAgent}
\SetKwFunction{FSyntax}{SyntaxParser}
\SetKwProg{Fn}{Function}{}{}

\Class{\FMain{$S$, $E$}}{
    \Fn{\textup{process}($Q$)}{
        code $C\gets$  \textup{generate\_code}([\textit{S}, \textit{E}, \textit{Q}])\;
        \For{function F \textup{in} C}{
            \If{F \textup{is not defined}}{
                $C^{\prime} \gets$ \FSub($S^\prime$, $E^\prime$).process($F$, context($F$))\;
                $C \gets$ \textup{append}([$C^{\prime}$, $C$])
            }
        }
        \Return $C$
    }
}

\Class{\FSub{$S^\prime$, $E^\prime$}}{
    \Fn{\textup{process(function \textit{F}, context \textit{CT})}}{
        query $Q^\prime \gets$ make\_query(\textit{F}, \textit{CT}) \;
        code $C\gets$  \textup{generate\_code}([\textit{S'}, \textit{E'}, \textit{Q'}])\;
        \For{function $F^\prime$ \textup{in} C}{
            \If{$F^\prime$ \textup{is not defined}}{
                $C^{\prime} \gets$ \FSub($S^\prime$, $E^\prime$).process($F^\prime$, context($F^\prime$))\;
                $C \gets$ \textup{append}([$C^{\prime}$, $C$])
            }
        }
        \Return $C$
    }
}

Code $\gets$ MainAgent($S$, $E$).process(\textit{Q})\;
\caption{Hierarchical Agent Code Generation}\label{alg:code-gen}
\end{algorithm}

% \subsubsection{Extra hint in task instruction}
% \textcolor{red}{Should we address the extra prompt for Gemini here?}

\subsection{Tool APIs design}
\label{sec:appendix_tools}
This section gives a detailed description of all the tools API that language models have access to. 

The Perception Tools are shown in Code Block \ref{code:perception tools}. The perception tools provide geometric information as function arguments for the motion planning tools, including object positions, object 3D bounding box, articulated joint information (moving direction or rotational axis), and plane surface normals. Perception tools are organized as object properties using object names as arguments and function names to specify the property type. With this \textbf{object-centric} design, LLMs only need to reason which object properties are involved in a specific task. Through our observation, LLMs are good at handling high-level task decomposition and context-consistent function calling rather than low-level geometric understanding or mathematical-consistent algorithm generation. This design makes the best use of LLM's strengths and avoids its weaknesses.  

% \TODO{Draw object-centric perception tools organization.}

These perception tools provide spatial parameters for trajectory generation and pick-and-place actions. For example, to generate a trajectory to open a door, the LLM needs to call the "\textit{get\_object\_joint\_info}" function to get its door joint axis, then to calculate the distance from the handle to the door joint, and further generate an arc path around this joint, with radius same as the distance from the door joint to handle, as demonstrated in "\textit{generate\_arc\_path\_around\_joint}." Then, the robot can call "\textit{follow\_path}" to execute this trajectory.

The Motion Planning Tools are shown in Code Block \ref{code:motion tools}. The motion planning tools are mostly re-wrappers of the Moveit Python move group interface. They hide system functional code under the hood, such as ROS synchronization and failure recovery behavior. Two API tools worth mentioning are:
\begin{itemize}
    \item \textit{grasp}. This tool executes a grasp motion with a given grasp pose: It first moves to a pre-grasp pose, as did in \cite{jiang2021synergies}\cite{fang2023anygrasp}, which is a pose retracted from the grasp pose on the approaching direction of the gripper. Assume the target grasp pose to be $\mat{P}=(\mat{R}|\vec{t})$, and the approaching direction is $\vec{[0,0,1]}$ under identity rotation, then the pre-grasp pose is $\mat{P^\prime}=(\mat{R}|\vec{t}-\mat{R}\cdot \vec{[0,0,0.1]})$. So this API plans and executes the path with two end-effector sub-goals $\{\mat{P^\prime}, \mat{P}\}$.
    \item \textit{parse\_place\_pose}. Although it may seem like a perception tool at first glance, this tool relies on the object bounding boxes, object mesh, and Moveit Inverse Kinematics solver to get the valid placing pose. Specifically, it first predicts a candidate object place center position by a heuristic algorithm given container bounding box and object bounding box and then solves a valid orientation for the gripper with Moviet IK. 
\end{itemize}

\begin{figure*}[t]
\begin{minipage}{\textwidth}
\centering
\begin{listing}[H]
\begin{minted}[tabsize=2, fontsize=\scriptsize, bgcolor=bg, breaklines, frame=lines,framesep=5pt]{python}
from perception_utils import (
    get_object_center_position,  # Returns the position of an object in the world frame. Returns: position: np.array [x,y,z]
    get_object_pose,             # Returns the pose of an object in the world frame. Returns: pose: Pose
    get_3d_bbox,                 # Returns the 3D bounding box of an object in the world frame. Args: object_name: str. Returns: bbox: np.array [x_min, y_min, z_min, x_max, y_max, z_max]
    get_object_name_list,        # Returns a list of names of objects present in the scene
    parse_adaptive_shape_grasp_pose, # Parse adaptive grasp pose for objects. Args: object_name: str, preferred_position: Optional(np.ndarray); preferred gripper tip point position; preferred_approach_direction: Optional(np.ndarray), preferred gripper approach direction; preferred_plane_normal: Optional(np.ndarray), preferred gripper plane normal direction. Returns: grasp_pose: Pose
    parse_central_lift_grasp_pose, # This method involves a vertical lifting action. The gripper closes at the center of the object and is not suitable for elongated objects and is not suitable for the objects with openings, as the gripper's width is really small. It is optimal for handling spherical and cuboid objects without any opening that are not ideal for off-center grasping. Args: object_name: str, description: Optional(str) in ['top', 'center'], Returns: grasp_pose: Pose
    detect_objects,              # Detect and update task-relevant objects' status in the environment. Call this function before interaction with environment objects. Args: object_list: Optional(List[str]), objects to detect.
    get_object_joint_info,       # Get the joint info of an object closest to a given position. Args: obj_name: str, name of the object; position: np.ndarray, select the joint closest to this position; type: str, allowed type of the joint, choice in ["any", "revolute", "prismatic"]. Returns: joint_info: dict, joint info of the object. {"joint_position":[x,y,z],"joint_axis":[rx,ry,rz],"type":str}
    get_plane_normal,            # Get the plane normal of an object closest to the given position. Args: obj_name: name of the object position: np.ndarray, select the plane closest to this position. Returns: np.ndarray, normal vector [x,y,z]
)
\end{minted}
\caption{\textbf{Perception tools} API in system prompt during code generation}
\label{code:perception tools}
\end{listing}
\end{minipage}
\end{figure*}

\begin{figure*}[t]
\begin{minipage}{\textwidth}
\centering
\begin{listing}[H]
\begin{minted}[tabsize=2, fontsize=\scriptsize, bgcolor=bg, breaklines, frame=lines,framesep=5pt]{python}
from motion_utils import (
    attach_object,  # Attaches an object to the robot gripper in the planning space. Call this function right after closing the gripper. Args: object_id: str. 
    detach_object,   # Detaches an object from the robot gripper in the planning space. Call this function right after opening the gripper. Args: object_id: str. 
    open_gripper,    # Open the gripper. No args.
    close_gripper,   # Close the gripper. No args.
    move_to_pose,    # Move the gripper to pose. Args: pose: Pose
    move_in_direction, # Move the gripper in the given direction in a straight line by certain distance. Note that you can use the surface normal and the joint axis. Args: axis: np.array, the move direction vector ; distance: float.
    generate_arc_path_around_joint, # Generate a rotational gripper path of poses around the revolute joint. Args: current_pose: Pose, current pose of the gripper; joint_axis: np.array, the joint axis of the revolute joint; joint_position: np.array, the joint position of the revolute joint; n: int, number of waypoints; angle: float, angle of rotation in degree. Returns: path: List[Pose]
    follow_path,     # Move the gripper to follow a path of poses. Args: path: List[Pose]
    get_gripper_pose, # Get the gripper pose. No args. Returns: pose: Pose
    grasp,           # Executes a grasp motion at the grasp_pose. Args: grasp_pose: Pose
    parse_place_pose, # Predict the place pose for an object relative to a receptacle. Args: object_name: str, receptacle_name: Optional(str), position: Optional(np.array) [x,y,z], . Returns: place_pose: Pose
)
\end{minted}
\caption{\textbf{Motion Planning tools} API in system prompt during code generation}
\label{code:motion tools}
\end{listing}
\end{minipage}
\end{figure*}

\subsection{Failed code case study}
\label{appendix: failed_code}
LLMs generate control code with a failure rate and all sorts of failure patterns, including grammar failures and logical fallacies. The grammar failures can be: 1) Hallucinated function named arguments, 2) wrong variable type and operation, 3) garbage code (usually when code is too long or the generation process goes out of control), etc. However, these are the basic code-generation capabilities. They are well-studied in a lot of general LLM code generation evaluation papers. Here, we discuss several failure cases due to the limitation of LLMs' reasoning capabilities on embodied tasks.  

\subsubsection{GPT-3.5-turbo}

\begin{itemize}
    \item \textbf{hard\_3, place the brown cup into an empty plate}
    Extra instruction in prompt: \textit{you need to find empty plate by checking if there is any object in the plate's 2D region on the table} 
    \\
    This script has the wrong assumption that if the min and max values of a plate's bounding box are zero, it is empty, which is obviously not correct.
\begin{minted}[tabsize=2, fontsize=\scriptsize, bgcolor=bg, breaklines, frame=lines,framesep=5pt]{python}
# Find an empty plate on the table
empty_plate = None
table_objects = get_object_name_list()
for object_name in table_objects:
	if object_name.endswith('_plate'):
		bbox = get_3d_bbox(object_name)
		if bbox[2] == 0 and bbox[5] == 0:  # Check if the plate is empty by checking its height (z-axis)
			empty_plate = object_name
			break
\end{minted}
    \item \textbf{hard\_4, clear the table by moving all table objects into the basket}
    \\
    Although it might look good initially, the script skips the table and basket. However, the function did return \textit{cabinet}, which the script does not need to pick. The master code is to construct a list of objects, only including objects that need to be picked and placed, rather than this exclusion logic. 
\begin{minted}[tabsize=2, fontsize=\scriptsize, bgcolor=bg, breaklines, frame=lines,framesep=5pt]{python}
# Iterate through each object on the table
for object_name in object_names:
    # Skip the table and the basket
    if object_name == 'table' or object_name == 'basket_with_fabric_liner':
        continue
    
    # Grasp the object
    open_gripper()
    grasp_pose = parse_adaptive_shape_grasp_pose(object_name)
    ...
\end{minted}

\end{itemize}

\subsubsection{Gemini}

\begin{itemize}
    \item \textbf{hard\_3, Place the brown cup into an empty plate}:
    Extra instruction in prompt: \textit{you need to find empty plate by checking if there is any object in the plate's 2D region on the table} 
    \\
    This task requires LLM to understand what is an "empty plate" with spatial reasoning process. However, Gemini generated a weird function to check if an object is in a plate by assuming objects have a name format \textit{plate\_apple}, which does not hold obviously. This failure can be attributed to Gemini's \textbf{lack of attention to instruction and hallucination}. 
% \begin{lstlisting}[ht]
% \begin{tcolorbox}
\begin{minted}[tabsize=2, fontsize=\scriptsize, bgcolor=bg, breaklines, frame=lines,framesep=5pt]{python}
def get_object_names_in_region(plate, region='table'):
  """
  Get a list of object names in a given region.

  Args:
    plate (str): The name of the plate.
    region (str): The name of the region.

  Returns:
    list: A list of object names in the region.
  """
  object_names = []
  for obj in get_object_name_list():
    # Check if the object is in the given region
    if region == 'table':
      if is_object_on_table(obj, plate):
        object_names.append(obj)
    elif region == 'receptacle':
      if get_receptacle_name_from_object_name(obj) == plate:
        object_names.append(obj)
  return object_names
\end{minted}

% # Detailed planning:
% # Step 1: Grasp the mug
% # Step 2: Move the mug to the top drawer
% # Step 3: Grasp the first drawer handle
% # Step 4: Open the first drawer
% # Step 5: Move the mug to the second drawer
% # Step 6: Release the mug
% # Step 7: Close the first drawer
% # Step 8: Release the first drawer handle
% \end{tcolorbox}
% \end{lstlisting}
\end{itemize}

\subsection{Task Data Generation and Dataset}
\label{subsec:task-generation}

One of the distinctive features of the RoboScript platform is the capability of task data generation automatically. The platform can generate task data of (task query, environment, generated code) to train custom embodied code generation models. This generation involves three steps:
\begin{enumerate}
    \item Randomly select candidate containers and pickable objects, sample an valid arrangement, and save to the environment definition file. 
    \item Collect table top arrangement information, convert to text description ("Apple is in the grey bowl. Banana is on the white plate."). Send the environment description to LLM, to prompt the LLM to suggest what tasks can be done for this environment. 
    \item For each task query, generate a normal code snippet to execute the task. 
\end{enumerate}

This approach ensures task diversity and complexity. We have already generated 100 randomly arranged tabletop environments and 4500+ grammatically correct code snippets and constructed into (task query, environment, generated code) tuples for some private projects. 

\begin{figure}
\centering
  \includegraphics[width=0.45\textwidth]{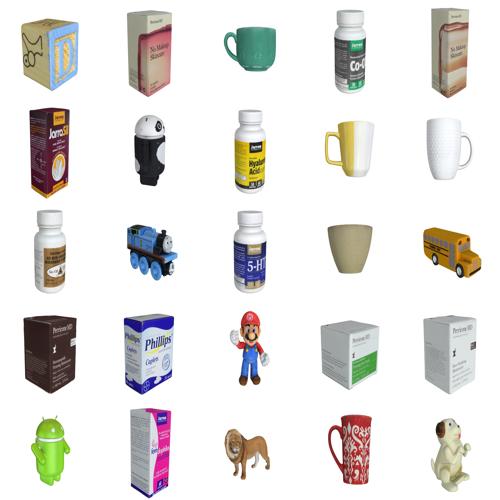} 
  \caption{\textbf{Pickable objects selected from Google Scanned Objects.}}
  \label{fig:pickable_objects}
\end{figure}

\begin{figure}
\centering
  \includegraphics[width=0.45\textwidth]{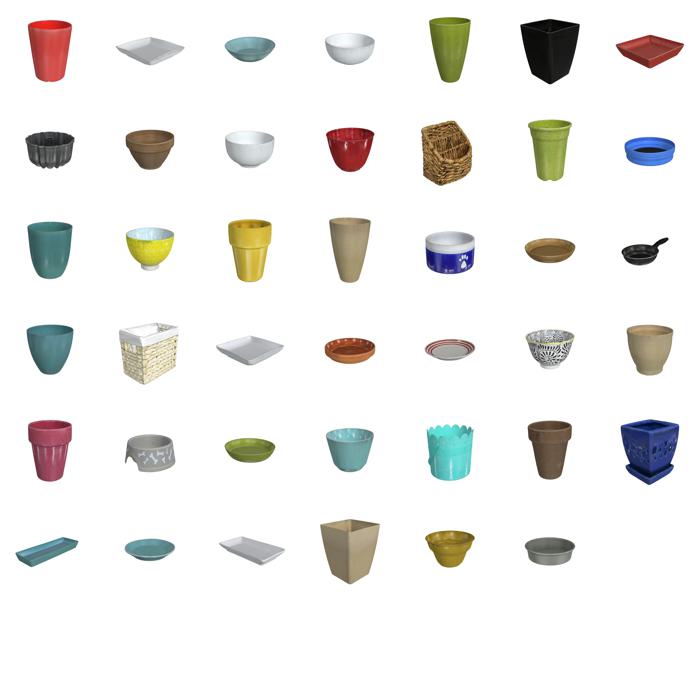} 
  \caption{\textbf{Container objects selected from Google Scanned Objects.}}
  \label{fig:container_objects}
\end{figure}

The models are selected from YCB or Google Scanned Objects. A collage of pickable objects and container objects is given in \figureautorefname \ref{fig:pickable_objects} and \figureautorefname \ref{fig:container_objects}. 

\begin{figure*}[htbp]
    \centering
    \begin{subfigure}[b]{0.23\textwidth}
        \centering
        \includegraphics[width=\textwidth]{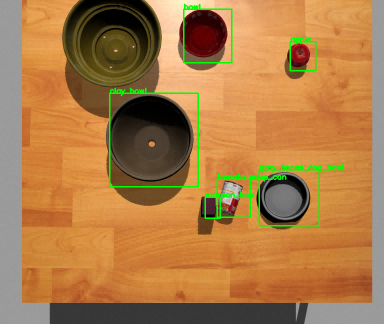}
    \end{subfigure}
    \begin{subfigure}[b]{0.23\textwidth}
        \centering
        \includegraphics[width=\textwidth]{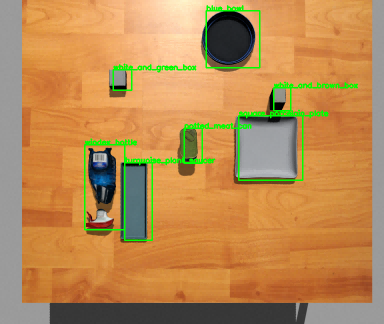}
    \end{subfigure}
    \begin{subfigure}[b]{0.23\textwidth}
        \centering
        \includegraphics[width=\textwidth]{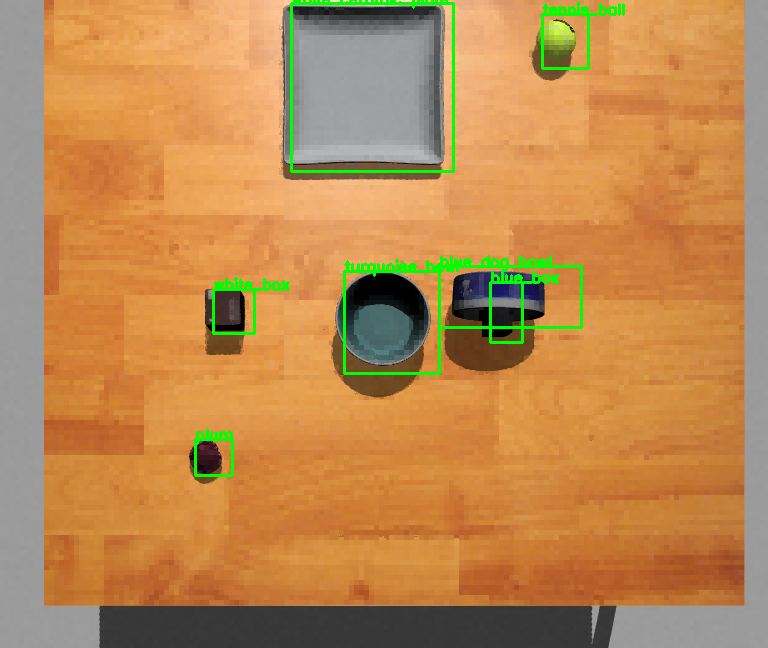}
    \end{subfigure}
    \begin{subfigure}[b]{0.23\textwidth}
        \centering
        \includegraphics[width=\textwidth]{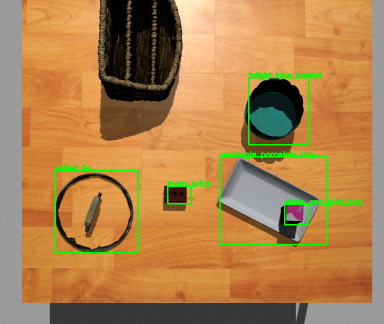}
    \end{subfigure}
    \\
    \begin{subfigure}[b]{0.23\textwidth}
        \centering
        \includegraphics[width=\textwidth]{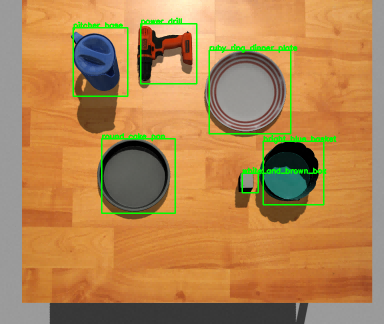}
    \end{subfigure}
    \begin{subfigure}[b]{0.23\textwidth}
        \centering
        \includegraphics[width=\textwidth]{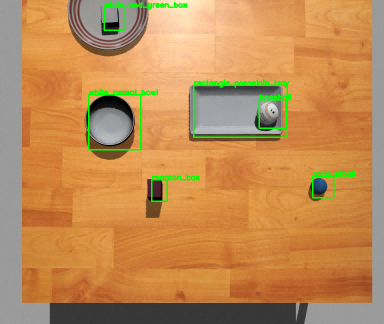}
    \end{subfigure}
    \begin{subfigure}[b]{0.23\textwidth}
        \centering
        \includegraphics[width=\textwidth]{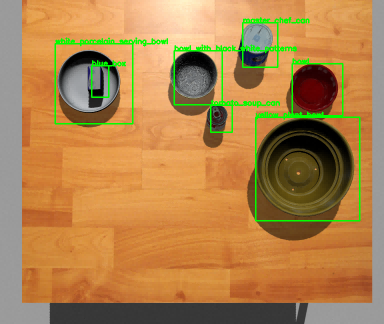}
    \end{subfigure}
        \begin{subfigure}[b]{0.23\textwidth}
        \centering
        \includegraphics[width=\textwidth]{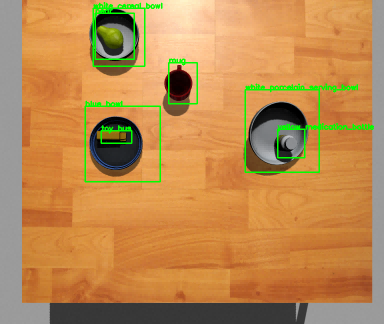}
    \end{subfigure}
    \caption{\textbf{Auto-generated Worlds.} We show the top view of the table and saved GT detection results of objects.}
    \label{fig:generated_worlds}
\end{figure*}

Some of the auto-generated worlds are shown in \figureautorefname \ref{fig:generated_worlds}. To better visualize its heuristic arrangments, we show the top view.

% \subsection{How should LLMs' choice of words align with multi-modal tools}
% In this section, we discuss one of the blind spot of previous literature, but could be a significant bottleneck of code generation for robot perception. The root for this issue can be summarized as the preference for object names of LLMs and multi-modal models could be different, which leads to poor grounding result.

\subsection{Implementation of~\algname{\alias}}\label{sec:implementation}

% Herein we discuss our instantiation of~\algname. We focus our discussions on design choices shared between the simulated and real-world domains. More details about the environment setup in each domain can be found in Appendix.

\textbf{LLMs and Prompting.}
% We follow prompting structure by~\citet{liang2022code}, which recursively calls LLMs using their own generated code, where each language model program (LMP) is responsible for a unique functionality (e.g., processing perception calls). We use GPT-4~\cite{openai2023gpt} from~\href{https://openai.com/api/}{OpenAI API}. Prompts are in Appendix.
We follow prompting structure by~\citet{liang2023code}, which recursively calls LLMs using their own generated code, where each language model program (LMP) is responsible for a unique functionality (e.g., processing perception calls). We use GPT-4~\cite{openai2023gpt} from~\href{https://openai.com/api/}{OpenAI API}.
For each LMP, we include 10 example queries and corresponding responses as part of the prompt.

% \textbf{VLMs and Perception.}
% Given an object/part query from LLMs, we first invoke open-vocab detector OWL-ViT~\cite{minderer2022simple} to obtain a bounding box, then feed it into Segment Anything~\cite{kirillov2023segment} to obtain a mask, and finally track the mask using video tracker XMEM~\cite{cheng2022xmem}. The tracked mask is used with RGB-D observation to reconstruct the object/part point cloud.

% \textbf{Value Map Composition.}
% We define the following types of value maps: affordance, avoidance, end-effector velocity, end-effector rotation, and gripper action. Each type uses a different LMP, which takes in an instruction and outputs a voxel map of shape $(100, 100, 100, k)$, where $k$ differs for each value map (e.g., $k=1$ for affordance and avoidance as it specifies cost, and $k=4$ for rotation as it specifies SO$(3)$). We apply Euclidean distance transform to affordance maps and Gaussian filters for avoidance maps. On top of value map LMPs, we define two high-level LMPs to orchestrate their behaviors: \texttt{planner} takes user instruction $\mathcal{L}$ as input (e.g., ``open drawer'') and outputs a sequence of sub-tasks $\ell_{1:N}$, and \texttt{composer} takes in sub-task $\ell_i$ and invokes relevant value map LMPs with detailed language parameterization.

\textbf{Motion Planning and control. }
We use MoveIt-integrated Open Motion Planning Library (OMPL), which is a collection of state-of-the-art sampling-based motion planning algorithms. Specifically, we use RRT (Rapidly-exploring Random Trees) and PRM (Probabilistic Roadmaps) by default for trajectory path sampling. 

For the robot control, by default, we use \textit{JointStateController} in the ros control package, taking as input the current joint states and joint goal states. It runs a PID controller at the backend at 50 Hz. The PID parameters are tuned and saved to individual Moveit robot configurations separately, mostly provided by each robot manufacturer. 
% We consider only affordance and avoidance maps in the planner optimization, which finds a sequence of collision-free end-effector positions $p_{1:N} \in \mathbb{R}^3$ using greedy search. Then we enforce other parametrization at each $p$ by the remaining value maps (e.g., rotation map, velocity map). The cost map used by the motion planner is computed as the negative of the weighted sum of normalized affordance and avoidance maps with weights $2$ and $1$. After a 6-DoF trajectory is synthesized, the first waypoint is executed, and then a new trajectory is re-planned at 5 Hz.

% \textbf{Environment Dynamics Model. }
% For tasks in which the specified ``entity of interest'' is the robot, we assume identity environment dynamics while replanning at every step to account for the latest observation. For tasks in which the ``entity of interest'' is an object, we study only a planar pushing model parametrized by contact point, push direction, and push distance. The heuristic-based dynamics model translates an input point cloud along the push direction by the push distance. We use MPC with random shooting to optimize for the action parameters. Then a pre-defined pushing primitive is executed based on the action parameters. However, we note that a primitive is not necessary when action parameters are defined over the end-effector or joint space of the robot, which would likely yield smoother trajectories but takes more time for optimization.
\textbf{Dynamics Model.}
We use the known robot dynamics model in all tasks, where it is used in motion planning for the end-effector to follow the waypoints. For the majority of our considered tasks where the ``entity of interest'' is the robot, no environment dynamics model is used (i.e., the scene is assumed to be static), but we replan at every step to account for the latest observation.
For tasks in which the ``entity of interest'' is an object, we study only a planar pushing model parametrized by contact point, push direction, and push distance. We use a heuristic-based dynamics model that translates an input point cloud along the push direction by the push distance.